\documentclass{article}
\usepackage[utf8]{inputenc}

\usepackage[utf8]{inputenc}
\usepackage{amsmath, amssymb, amsthm}
\usepackage{upgreek}
\usepackage{xcolor}
\usepackage{fullpage}
\usepackage{caption}
\usepackage{graphicx}
\usepackage{enumitem}
\usepackage{subcaption}
\usepackage{multirow}
\usepackage{authblk}
\usepackage{fancyhdr}

\DeclareSymbolFont{AMSb}{U}{msb}{m}{n}
\DeclareMathSymbol{\N}{\mathbin}{AMSb}{"4E}
\DeclareMathSymbol{\Z}{\mathbin}{AMSb}{"5A}
\DeclareMathSymbol{\R}{\mathbin}{AMSb}{"52}
\DeclareMathSymbol{\Q}{\mathbin}{AMSb}{"51}
\DeclareMathSymbol{\I}{\mathbin}{AMSb}{"49}
\DeclareMathSymbol{\C}{\mathbin}{AMSb}{"43}
\DeclareMathSymbol{\F}{\mathbin}{AMSb}{"46}

\newcommand{\grad}{\textnormal{grad }}

\theoremstyle{remark}

\fancypagestyle{release}{\fancyhf{}\fancyfoot[R]{Approved for public release; distribution is unlimited.}}

\title{On-Manifold Projected Gradient Descent}
\author[$\dagger$]{Aaron Mahler}
\author[$\ddagger$]{Tyrus Berry}
\author[$\dagger$]{Tom Stephens}
\author[$\ddagger$]{Harbir Antil}
\author[$\dagger$]{Michael Merritt}
\author[$\ddagger$]{Jeanie Schreiber}
\author[$\star$]{Ioannis Kevrekidis}
\affil[$\dagger$]{Teledyne Scientific \& Imaging, LLC}
\affil[$\ddagger$]{Center for Mathematics and Artificial Intelligence\\ George Mason University}
\affil[$\star$]{Departments of Chemical and Biomolecular Engineering \& Applied Mathematics and
Statistics, Johns Hopkins University}
\date{\today}

\begin{document}

\maketitle
\thispagestyle{release}

\begin{abstract}
%% submitted abstract:
This work provides a computable, direct, and mathematically rigorous approximation to the differential geometry of class manifolds for high-dimensional data, along with nonlinear projections from input space onto these class manifolds. 
The tools are applied to the setting of neural network image classifiers, where we generate novel, on-manifold data samples, and implement a projected gradient descent algorithm for on-manifold adversarial training.
The susceptibility of neural networks (NNs) to adversarial attack highlights the brittle nature of NN decision boundaries in input space. 
Introducing adversarial examples during training has been shown to reduce the susceptibility of NNs to adversarial attack;
however, it has also been shown to reduce the accuracy of the classifier if the examples are not valid examples for that class.
Realistic ``on-manifold" examples have been previously generated from class manifolds in the latent of an autoencoder.
Our work explores these phenomena in a geometric and computational setting that is much closer to the raw, high-dimensional input space than can be provided by VAE or other black box dimensionality reductions.  
We employ conformally invariant diffusion maps (CIDM) to approximate class manifolds in diffusion coordinates, and develop the Nystr\"{o}m projection to project novel points onto class manifolds in this setting.
On top of the manifold approximation, we leverage the spectral exterior calculus (SEC) to determine geometric quantities such as tangent vectors of the manifold. 
% Additionally, SEC provides a path to overlay semantic directions on these class manifolds. 
We use these tools to obtain adversarial examples that reside on a class manifold, yet fool a classifier.
These misclassifications then become explainable in terms of human-understandable manipulations within the data, by expressing the on-manifold adversary in the semantic basis on the manifold.
\end{abstract}

\section{Introduction} \label{sec:introduction}
% \subsection{Background} \label{sec:background}
% Can we talk more about semantic label Nystrom to make it clearer that we're explaining misclasses wrt semantic labels
Despite their superior performance at image recognition, neural network (NN) classifiers are susceptible to adversarial attack, and their performance can degrade significantly with small perturbations to the input \cite{szegedy_intriguing_2014, tabacof_exploring_2016, moosavi-dezfooli_universal_2017}. 
The brittle performance of NNs when given novel inputs can be attributed to their intricate high-dimensional decision boundaries, which fail to generalize robustly outside of the training data. 
This problem is epitomized by the observation that NNs are excellent interpolators but poor extrapolaters. 

% Go into a brief history of adversarial attacks, FGSM, single pixel,
Crafting attacks to deceive NNs with minimal changes to the input has been shown to be remarkably easy when the attacker has full access to the NN architecture and weights. 
The fast gradient sign method is one of the earliest attack methods that crafts adversarial examples by taking the sign of the gradient of the loss function in order to perturb the input in the direction that maximizes the loss in pixel space \cite{goodfellow_explaining_2015}.
Other methods take a number of smaller steps in directions to find the smallest perturbation required to misclassify an input \cite{bengio_adversarial_2018, moosavi-dezfooli_deepfool_2016, carlini_towards_2017, madry_towards_2019}. 
Most of these methods use the gradient of the NN loss function for a given input as a way to determine directions of maximal confusion, i.e. directions leading to the closest decision boundary in the high-dimensional pixel space.
There also exists single-pixel attacks that use differential evolution with no gradient information and are able to reliably fool NNs \cite{su_one_2019}.

% Go into methods for adversarial defense training
% -Adversarial training, defensive distillation, gradient masking, ensemble methods, input preprocessing, 
Various methods have been proposed to make NNs more robust to adversarial attack.
Adversarial training is a common choice because it involves using attack inputs as additional training data, 
thereby allowing the NN decision boundary to more correctly classify that data. 
Commonly the gradient of the NN will be used to augment the dataset for this purpose \cite{goodfellow_explaining_2015, madry_towards_2019}.
On the other hand, gradient masking is a method that attempts to create a network that does not have useful information in the gradient, so that it cannot be exploited for creating attacks \cite{papernot_practical_2017}. 
These types of networks have found to still be vulnerable though to similar attacks that work on NNs with useful gradients \cite{athalye_obfuscated_2018, papernot_practical_2017}.
Defensive distillation is a gradient-free method that uses two networks where the second network is trained using the distilled (softened) outputs of the first network \cite{papernot_practical_2017}. 
Training on distilled outputs is done to create less irregular decision boundaries, which in turn results in being less prone to misclassifying small perturbations. 
Ensemble methods use the output of multiple models, which results in less effective attacks since it is unlikely the models are sensitive to the exact same attacks \cite{tramer_ensemble_2020}.
Input preprocessing can also be applied to try to mitigate or remove adversarial perturbations. 
This can be done in a model agnostic way such as filtering or compressing the data \cite{yin_war_2020}, detecting adverasial inputs with feature squeezing \cite{xu_feature_2018}, or using an autoencoder to denoise the input \cite{cho_dapas_2020}. 

% Paragraph on robustness is at odds with accuracy
One popular method of adversarial training uses small steps along the network gradient that are only allowed to step so far away from the original input, called projected gradient descent (PGD) \cite{madry_towards_2019}. 
The dataset is augmented with examples that are maximally confusing to the NN during training, but the augmented data points are only allowed to be $\epsilon$ far away from true data points. 
This results in a marked improvement to the NN when it is attacked with perturbations of the same strength. 
However, PGD trained networks show a decrease in accuracy on clean inputs and the accuracy goes down as the size of the $\epsilon$-ball allowed for augmentations increases \cite{engstrom_robustness_2019}. 
The degradation of accuracy and the rise of robustness could be due to several factors, such as overfitting the model to adversarial examples, or from adversarial examples that are not actually representative of the class of the input that was perturbed.
The the trade off between robustness and accuracy has been noted to occur with many flavors of adversarial training and it has even been conjectured that robustness and accuracy may be opposed to each other for certain NNs \cite{tsipras_robustness_2019, su_is_2018}.

% Detail the on manifold adversary view point
On the other hand, it has also been shown that in some cases a more careful choice of adversarial examples can create robust NNs that are also on-par with standard networks at generalizing to unseen validation data \cite{stutz_disentangling_2019}. 
This was explained by the fact that adversarial training such as PGD creates samples that are not truly on the manifold of that data's class label. 
The NN is then tasked with learning a decision boundary for the training data as well as randomly noisy data, resulting in the compromise between accuracy and robustness for those types of adversarial training. 
In \cite{stutz_disentangling_2019}, they found perturbations in a latent space learned from the training data. 
Perturbing in an $\epsilon$ ball in the latent space was surmised to be on a class manifold and therefore a new augmentation that was representative of that class. 
Adversarial training in a way that is agnostic to the underlying geometry of the data itself therefore seems to be a root cause for the trade-off between robustness and accuracy.

The above mitigations to adversarial attacks all proceed from the perspective that the neural network has simply not been fed enough variation in the collected training data to learn decision boundaries that adhere to the full underlying data manifolds.
From this perspective, injecting adversarial examples into the original training data ``pushes out" incursions by the decision boundary into the true manifold. 
An alternative perspective is that adversarial examples do not result from so-called bugs in the decision boundaries, but are instead features of the data \cite{ilyas_adversarial_2019}.
From the features perspective, adversarial training is a data cleaning process -- the original data has pixel correlations across classes that our eyes cannot detect, and computed adversarial perturbations act to wash those away.
While we do not embark on our applications from this perspective, the mathematical tools developed here are ideally suited for extending their hypotheses and result.

% Manifold hypothesis, one sentence summary of our goal and segue to next section
Our application of on-manifold adversarial training connects the learning problem to the manifold hypothesis and manifold modeling techniques. 
For natural images, the manifold hypothesis suggests that the pixels that encode an image of an object, together with the pixel-level manipulations that transform the scene through its natural, within-class variations (rotations, articulations, etc), organize along class manifolds in input space.
In other words, out of all possible images drawn from an input space, while the vast majority look like random noise, the collection of images that encode a recognizable object (a tree, a cat, or an ocean shoreline) are incredibly rare, and the manifold hypothesis claims that those images should be distributed throughout input space along some coherent geometric structure.
On-manifold adversarial training aims for a NN to better capture the underlying structure in the data. 
% In this work we use novel manifold modeling techniques that do not rely on autoencoders and show how to create on-manifold adversarial examples with the aim of creating veridical NNs whose decision boundaries are more closely aligned with reality. 
In this work we use novel manifold modeling techniques that do not rely on autoencoders or black box neural networks. We demonstrate creating on-manifold adversarial examples that are explainable in terms of their semantic labels.

\subsection{Manifold Learning and CIDM} \label{sec:manifold_learning}

Manifold learning emerged as an explanation for how kernel methods were able to perform regressions and identify low-dimensional coordinates from much higher dimensional data sets than would be be possible according to normal statistics.  Assuming that the data was actually lying on a submanifold of the data space, it appeared that the kernel methods (kernel regression, kernel PCA, etc.) we able to leverage this intrinsically lower dimensional structure.  

The first advance in understanding this effect rigorously was Laplacian Eigenmaps \cite{eigenmaps}. They employed a Gaussian kernel to build a complete weighted graph on the data set with weights $K_{ij}=k(x_i,x_j) = \exp(-||x_i-x_j||^2/(2\epsilon^2))$. This was a very common choice of radial basis kernel at the time and a natural first choice for analysis.  Laplacian Eigenmaps then constructs the weighted graph Laplacian $L=\frac{D-K}{\epsilon^2}$ where diagonal matrix $D_{ii} = \sum_j K_{ij}$ is called the degree matrix.  In the limit as the number of data points goes to infinity, the Laplacian matrices become larger and larger, and if the bandwidth, $\epsilon$, is taken to zero at an appropriate rate this sequence of matrices are shown to converge to the Laplace-Beltrami operator on the manifold that the data were sampled from.  This was the first rigorous connection between the somewhat ad hoc kernel methods and the intrinsic geometry of the data. 

Unfortunately, the assumptions required to prove the key theorem of Laplacian Eigenmaps were overly restrictive in practical settings.  In addition to only applying to a single kernel function (when in practice many different kernel functions were known empirically to have similar behavior), Laplacian Eigenmaps also required the data to be sampled uniformly from the underlying manifold.  This is a somewhat technical requirement, a embedded manifold (such as the one the data are assumed to lie on) inherits a natural measure from the ambient data space which is called the \emph{volume form}.  We can think of this volume form as a distribution, and when the data are sampled from this natural distribution it is called \emph{uniformly} sampled.  However, there is no reason for the data to have been be uniformly collected in this sense.  For example, if your data lies on a unit circle, there is no reason that the data could not be more densely collected on one side of the circle and more sparsely collected on the other side, but Laplacian Eigenmaps did not allow for this in their theorem.  These restrictions meant that the applicability of kernel methods to resolving the intrinsic geometry of a real dataset was still seen as rather tenuous.

Diffusion Maps \cite{diffusion} resolved these concerns and solidified the connection between a large class of kernel methods and the intrinsic geometry of the data.  The idea of Diffusion Maps turns out to be fairly simple although the technical details of the theorems are somewhat challenging. The key idea is that the degree matrix, $D_{ii} = \sum_j k(x_i,x_j)$ is actually a classical kernel density estimator, meaning that if the data is not sampled uniformly then $D_{ii}$ will be proportional to the true sampling density (up to higher order terms in $\epsilon$ which can be carefully accounted for).  Diffusion Maps begins by generalizing the kernel density estimation results to data sampled on manifolds, and then uses the estimated density to de-bias the kernel.  De-biasing the kernel turns out to be a simple normalization procedure called the \emph{diffusion maps normalization} which constructs the normalized kernel,
\[ \hat K = D^{-1}KD^{-1} \]
and then recomputes the new degree matrix $\hat D_{ii} = \sum_j \hat K_{ij}$ and finally the Diffusion Maps graph Laplacian $\hat L = \frac{\hat D - \hat K}{\epsilon^2}$. The Diffusion Maps theorems showed that for any radial basis kernel that had exponential decay in distance, and for data collected from any smooth distribution on the manifold, their new graph Laplacian, $\hat L$, converged to the Laplace-Beltrami operator on the manifold.  Moreover, the Diffusion Maps theorems also showed (although this was only realized in later works, e.g. \cite{berry2016local}) that even when their normalization was not used, the classical graph Laplacian converged to a Laplace-Beltrami operator with respect to a conformal change of metric.  This ultimately showed that all kernel methods with radial basis kernels that had fast decay were finding the intrinsic geometry of the data (possibly up to a change of measure).  Later works would generalize the Diffusion Maps theorems to include all kernels that had sufficiently fast decay in the distance between points (so not just radial basis functions) \cite{berry2016local}.  

At this point, we should address why both Laplacian Eigenmaps and Diffusion Maps have the word ``Maps" in them.  This goes back to the motivation that was driving the development of these new theories.  In particular, both methods were motivated by Kernel PCA, which interpreted the eigenvectors of the kernel matrix as providing the coordinates of a mapping into a new space, often called a `feature space'.  Ironically, this mapping interpretation arose from the theory of Reproducing Kernel Hilbert Spaces, where the kernel induces a map into a \emph{function} space (not a Euclidean space).  However, since the kernel matrix, $K$, has as many rows and columns as there were data points, the eigenvectors of the kernel matrix have as many entries as there are data points, so inevitably these were visualized and interpreted as new coordinates.  Diffusion Maps and Laplacian Eigenmaps were trying to show that this `mapping' preserved intrinsic aspects of the geometry while also reducing dimension, and while the first part is partially correct, the dimensionality reduction aspect of the Diffusion Maps turns out to not be guaranteed.  However, this was merely a case of applying the wrong interpretation to the results.  In fact what Diffusion Maps had proven was much better than any fact about a mapping.  By recovering the Laplace-Beltrami operator on the manifold, and its eigenfunctions, Diffusion Maps unlocked the door and allowed access to every single aspect of the geometry of the data.  Moreover, the eigenfunctions provide a generalized Fourier basis for analysis of functions and operators on the data set, and have been used in regression, interpolation, forecasting, filtering, and control applications.

In order to leverage the opening that Diffusion Maps has created to learning manifold geometry from data, we will need several recent advances that improve and apply the original theory.  First, it turns out that for real data sets in high dimensions, the fixed bandwidth kernels discussed so far have difficulty adjusting to large variations in sampling density. To compensate for this a variable bandwidth kernel is needed \cite{berry2016}, which can automatically adjust to have a small bandwidth and high resolution in areas of data space that are densely sampled, while keeping a large bandwidth and a lower resolution representation of sparsely sampled regions of data space.  The ultimate evolution of the variable bandwidth kernels is the \emph{Conformally Invariant Diffusion Map} (CIDM) \cite{berry2016local,berry2019consistent} which we introduce in Section \ref{sec:cidm}.

The next tool we will need is a rigorous method for extending/interpolating all of the discrete representations of functions, mappings, and operators to be able to operate on any new input data.  Here we use a regularized version of a standard method called the Nystr\"{o}m extension, introduced in Section \ref{sec:nystrom_background}. Although this basic method of interpolation is well established, we will apply it in ways that have never been considered before to achieve powerful new methods and results.

Finally, we mentioned above that the Laplace-Beltrami operator unlocks the door to access all the hidden geometry of the data.  This is due to a technical result which says that if you know the Laplace-Beltrami operator you can recover the Riemannian metric on the manifold, and the Riemannian metric completely determines all aspects of the geometry (from dimension and volume to curvature to geodesics and everything in between).  However, until recently this was a purely abstract possibility, and there was no actual method for constructing these geometric quantities starting from the Laplace-Beltrami operator.  This was achieved in 2020 with the creation of the Spectral Exterior Calculus (SEC), which re-builds all of differential geometry starting just from the Laplace-Beltrami operator.  While we will not require every aspect of the SEC here, the basic philosophy of its construction will be fundamental to the way that we will construct vector fields and in a particular tangent vectors on the manifold, so a brief introduction will be given in Section \ref{sec:secbackground}.

\subsubsection{Conformally Invariant Diffusion Map (CIDM)}\label{sec:cidm}

As mentioned above, the original version of Diffusion Maps uses a fixed bandwidth kernel of the form $J(x,y) = h(||x-y||^2/\epsilon^2)$.  Here $h$ is called the shape function and is assumed to decay quickly to zero as the input (distance) goes to infinity.  A typical choice for $h$ is the exponential function $h(z) = \exp(-z)$, so moderate differences in distances leads to large difference in the values of $h$.  This becomes particularly problematic in terms of the distance to the nearest neighbors.  If the distances from a data point to its nearest neighbors are large (relative to the bandwidth $\epsilon$) then the values of the kernel become very close to zero. This means that even though our weighted graph is technically still connected, the weights are so close to zero that it becomes numerically disconnected, which causes the Diffusion Map to interpret such data points as disconnected from the rest of the data set. On the other hand, we want the kernel function to decay quickly beyond the nearest neighbors to localize the analysis and make the resulting kernel matrix approximately sparse.

 When the density of points varies widely, it becomes very difficult to find a single bandwidth parameter $\epsilon$ which achieves these two goals across the data set.  One tends to have to choose the bandwidth large enough to connect with the sparsest region of data, and this large bandwidth value results in a loss of resolution in the denser sampled regions.  This trade-off is examined rigorously in \cite{berry2016} which introduces variable bandwidth kernels and generalizes the diffusion maps expansions for such kernels.  The best practical implementation of a variable bandwidth approach was the introduced in \cite{berry2019consistent} which is a variable bandwidth version of the Conformally Invariant Diffusion Map (CIDM) that was introduced in \cite{berry2016local}.

The CIDM starts by re-scaling the distance using the distances to the nearest neighbors, namely,
\begin{equation}\label{cidm_distance}
\delta(x,y) \equiv \frac{d(x,y)}{\sqrt{d(x,\textup{kNN}(x))d(y,\textup{kNN}(y))}}
\end{equation}
where $\textup{kNN}$ returns the $k$-th nearest neighbor of the input point from the training data set.  Note that the distance to the $k$-th nearest neighbor is a consistent estimator of the density to the power of $-1/d$ where $d$ is the dimension of the manifold.  Thus, when the local density is high, the distance to the kNN will be small, and conversely when the local density is sparse the distance to the kNN will be large.  Thus, $\delta(x,y)$ has  re-scaled the distances into a unit-less quantity which will be on the same order of magnitude for the $k$-th nearest neighbors of all the data points.   

Inside the kernel we will use the square of this quantity,
\[ \delta(x,y)^2 = \frac{d(x,y)^2}{d(x,\textup{kNN}(x))d(y,\textup{kNN}(y))} \]
which is also more convenient for derivatives. Next we use the dissimilarity $\delta$ in a kernel,
\[ k(x,y) \equiv h(\delta(x,y)^2/\epsilon^2) \]
where $\epsilon$ is a global bandwidth parameter and $h:[0,\infty)\to [0,\infty)$ is a called a shape function (examples include $h(z) = e^{-z}$ 
as mentioned above or even simply the indicator function $h(z) = 1_{[0,1]}(z)$).  We can then build the kernel matrix $K_{ij} = k(x_i,x_j)$ on the training data set, and the diagonal degree matrix $D_{ii} = \sum_j K_{ij}$ and the normalized graph Laplacian $L \equiv I-D^{-1}K$.  

We should note that in \cite{berry2019consistent}, it was shown that, uniquely for the CIDM, the unnormalized Laplacian $L_{\rm un} \equiv D-K$ has the same limit as the normalized Laplacian in the limit of large data, however, the normalized Laplacian, $L$, has some numerical advantages.  Numerically it is advantageous to maintain the symmetry of the problem by finding the eigenvectors of the similar matrix, $L_{\rm sym} \equiv D^{1/2}LD^{-1/2} = I - D^{-1/2}KD^{-1/2}$.  Finally, we are interested in the smoothest functions on the manifold, which are the minimizers of the energy defined by $L$, however it is easier to find the largest eigenvalues of $K_{\rm sym} \equiv D^{-1/2}KD^{-1/2}$. (Recall that maximal eigenvalues can be found with power iteration methods which are fast than the inverse power iterations required for finding smallest eigenvalues).  Once we have computed the eigenvectors $K_{\rm sym}\vec v = \lambda \vec v$, then it is easy to see that $\vec \phi = D^{-1/2}\vec v$ are eigenvectors $D^{-1}K$ with the same eigenvalues, and $\vec\phi$ are also eigenvectors of $L$ with eigenvalues $\xi = 1-\lambda$.  Thus when $\lambda$ are the largest eigenvalues of $K_{\rm sym}$, the corresponding $\xi$ will be the smallest eigenvalues of $L$.

 We will refer to $L$ as the CIDM Laplacian, and we will use the eigenvectors and eigenvalues of $L$ to represent the geometry of the data manifold. The eigenvectors of the CIDM Laplacian, $L \vec \phi = \lambda \vec \phi$ are vectors of the same length as the data set, so the entries of these eigenvectors are often interpreted as the values of a function on the data set, namely $\phi(x_i) = \vec \phi_i$.  Of course, we have not really defined a function $\phi$ since we have only specified its values on the data set.  However, in the next section we will show how to define a function $\phi$ on the whole data space that takes the specified values on data set. This method is called the \emph{Nystr\"{o}m extension} because it extends the function from the training data set to the entire data space.

In \cite{berry2016local} the CIDM Laplacian, $L$, was shown to converge (in the limit of infinite data and bandwidth going to zero) to the Laplace-Beltrami operator of the hidden manifold with respect to a conformal change of metric that has volume form given by the sampling density.  The Laplace-Beltrami operator encodes all the information about the geometry of the manifold (see Appendix \ref{sec} for details) which is why methods such as Diffusion Maps and the CIDM are called \emph{manifold learning} methods.  Moreover, it was shown in \cite{berry2016, berry2016local, berry2019consistent} that the CIDM construction using the $k$-nearest neighbors density estimator as described above, does not require the so-called `Diffusion Maps normalization'.  The CIDM gives the unique choice of conformal geometry for which a standard unnormalized graph Laplacian is a consistent estimator of a Laplace-Beltrami operator \cite{berry2019consistent}.  Empirically we have found that this variable bandwidth kernel construction is much more robust to wide variations in sampling density.

We should note that in the Nystr\"{o}m extension section below we will make use of the following normalized kernel,
\[ \hat k(x,y) = \frac{k(x,y)}{\sum_{i=1}^N k(x,x_i)} = \frac{h(\delta(x,y)^2/\epsilon^2)}{\sum_{i=1}^N h(\delta(x,x_i)^2/\epsilon^2)} \]
since this corresponds to $D^{-1}K$ as discussed above (namely if $K = k(x_i,x_j)$ then $\hat k(x_i,x_j) = (D^{-1}K)_{ij}$).
Finally, in order to reduce sensitivity, we often use the average of the distances to the k-nearest neighbors in the re-scaling so the dissimilarity would then be,
\[ \delta(x,y)^2 = \frac{d(x,y)^2}{\frac{1}{k}\sum_{i=1}^k d(x,\textup{iNN}(x)) \frac{1}{k}\sum_{j=1}^k d(y,\textup{jNN}(y))} \]
where iNN refers to the $i$-th nearest neighbor so the summations are averaging the distances to the $k$-nearest neighbors.

\subsubsection{Nystr\"{o}m Extension: Interpolation and Regularization}\label{sec:nystrom_background}

In this Section we introduce the Nystr\"{o}m extension, which is the standard approach for extending diffusion maps eigenfunctions (and thus the ``diffusion map") to new data points.  Once the eigenfunctions can be extended, arbitrary functions can also be extended by representing them in the basis of eigenfunctions; this approach can be used to extend any sufficiently smooth function to new data points in input space.  Since in practice we can only represent a function with finitely many eigenfunctions, the truncation onto this finite set gives us a regularized, or smoothed, regression.

Given an eigenvector $K\vec\phi = \lambda\vec\phi$ of a kernel matrix $K_{ij} = k(x_i,x_j)$, we can extend the eigenvector to the entire input space by,
\begin{equation}\label{nystromext} \phi(x) \equiv \frac{1}{\lambda}\sum_{j=1}^N k(x,x_j)(\vec\phi)_j \end{equation}
which is called the Nystr\"{o}m extension.  Note that here $k$ is an abstract kernel which may incorporate CIDM normalizations inside the shape function as well as normalization such as the Diffusion Maps normalization and/or Markov normalization outside of the shape function. 
 For example, for CIDM, the Nystr\"{o}m extension of an eigenfunctions is,
\[ \phi(x) = \frac{1}{\lambda}\sum_{j=1}^N \hat k(x,x_j)\phi(x_j) = \frac{\sum_{j=1}^N h(\delta(x,x_j)^2/\epsilon^2)\phi(x_j)}{\lambda \sum_{i=1}^N h(\delta(x,x_i)^2/\epsilon^2)} \] 
where the CIDM kernel $\hat k$ takes the place of the abstract kernel $k$ in \eqref{nystromext}.  Notice that evaluating $\hat k$ involves computing the dissimilarity $\delta$ between and arbitrary point $x$ and a training data point $x_j$, which in turn requires finding the $k$ nearest neighbors of the point $x$ from the training data set.  Thus, evaluating the abstract kernel $k$ may actually depend on the entire training data set, however, in this section we will consider the training data set as fixed and treat its influence on $k$ as hidden parameters that define the kernel $k$.  We should note that although everything in this section can be applied to any kernel, a simple radial basis function kernel with no normalizations and a fixed bandwidth has fairly poor performance for the off-manifold extensions we will discuss later in this section.

\begin{figure}[!h]
    \centering
    \includegraphics[width=0.24\linewidth]{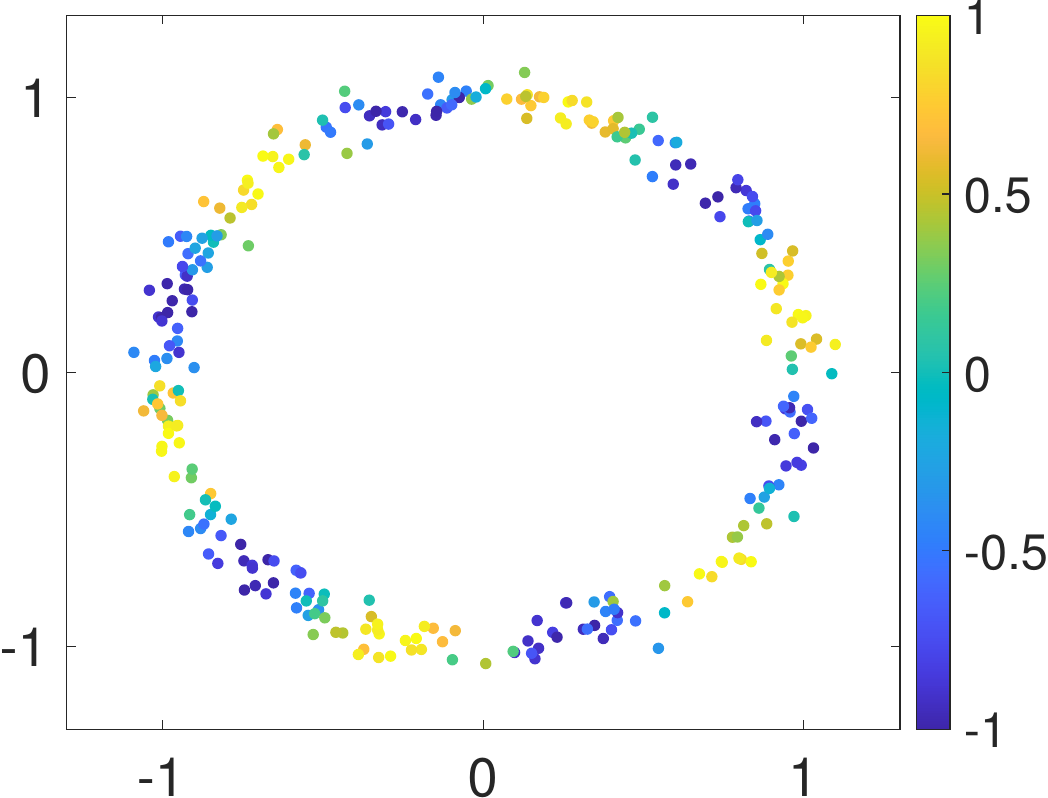}\includegraphics[width=0.24\linewidth]{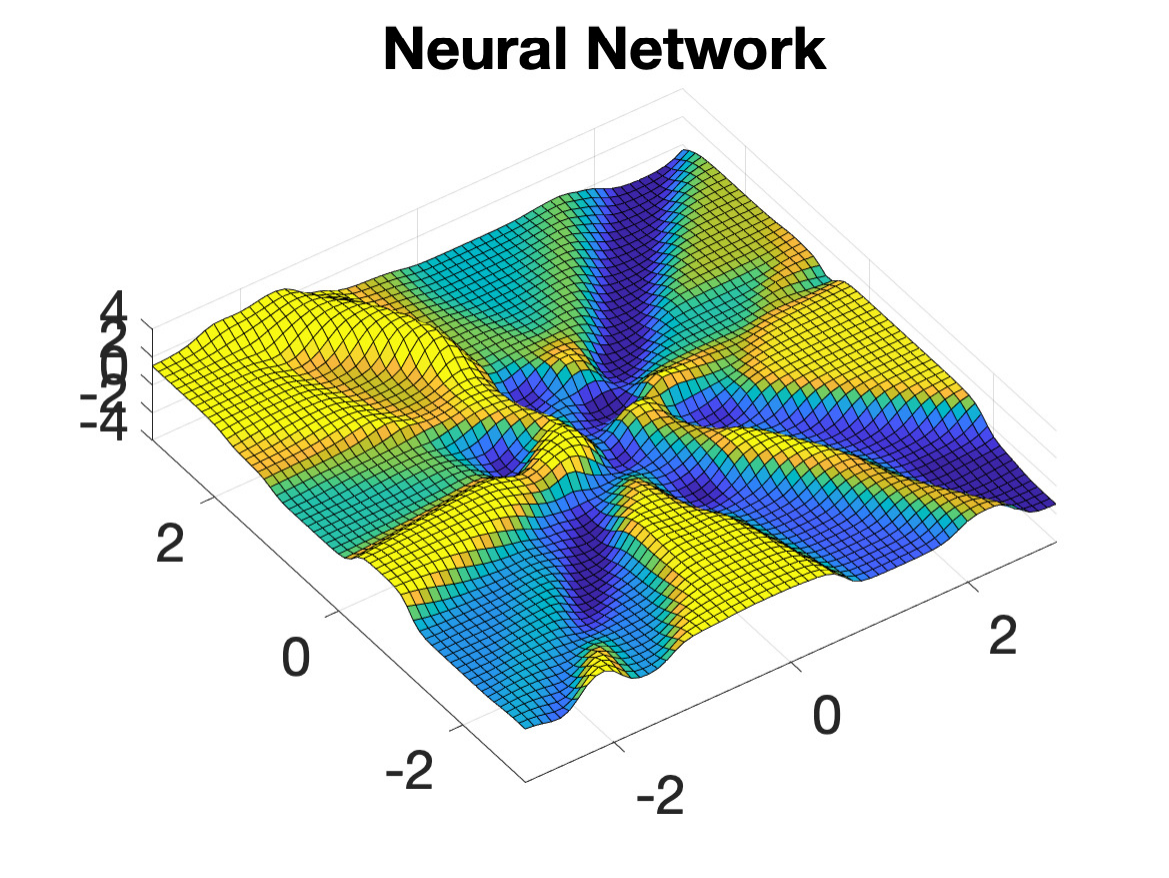}
    \includegraphics[width=0.24\linewidth]{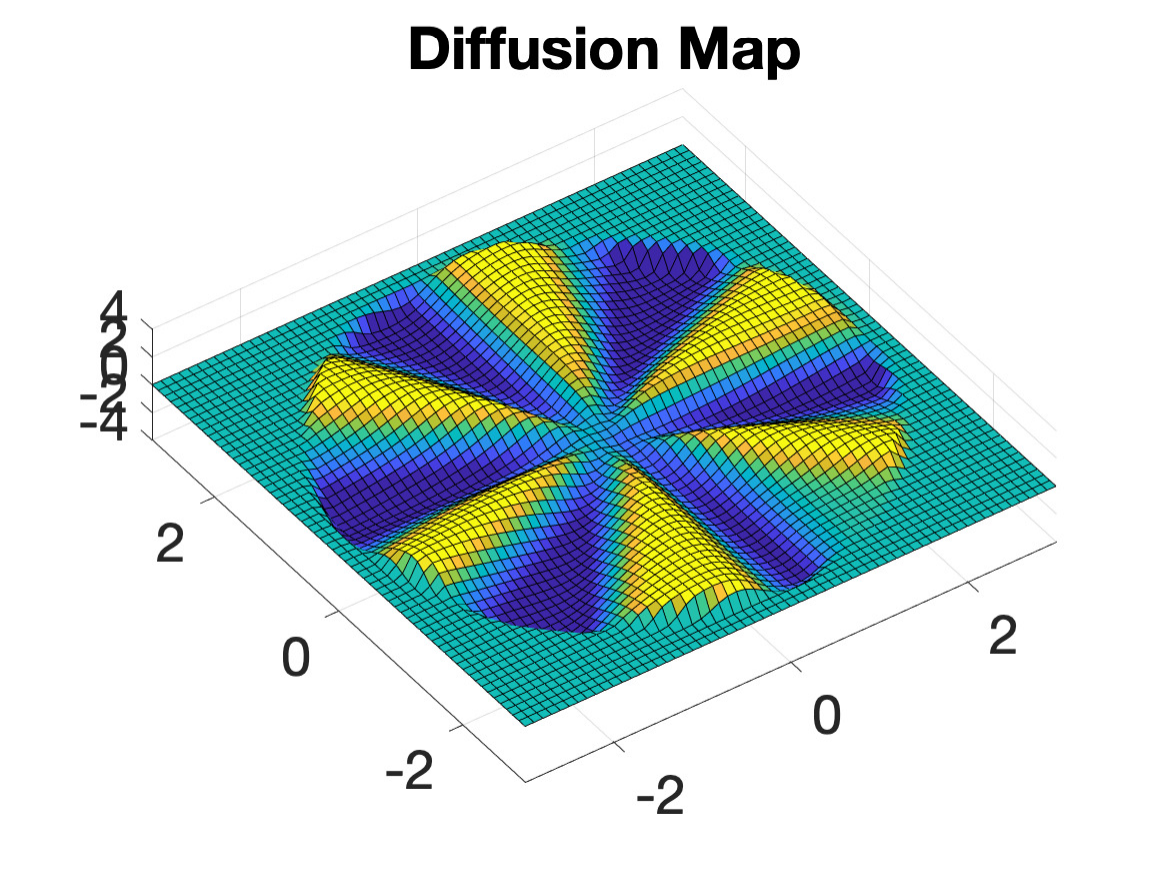}\includegraphics[width=0.24\linewidth]{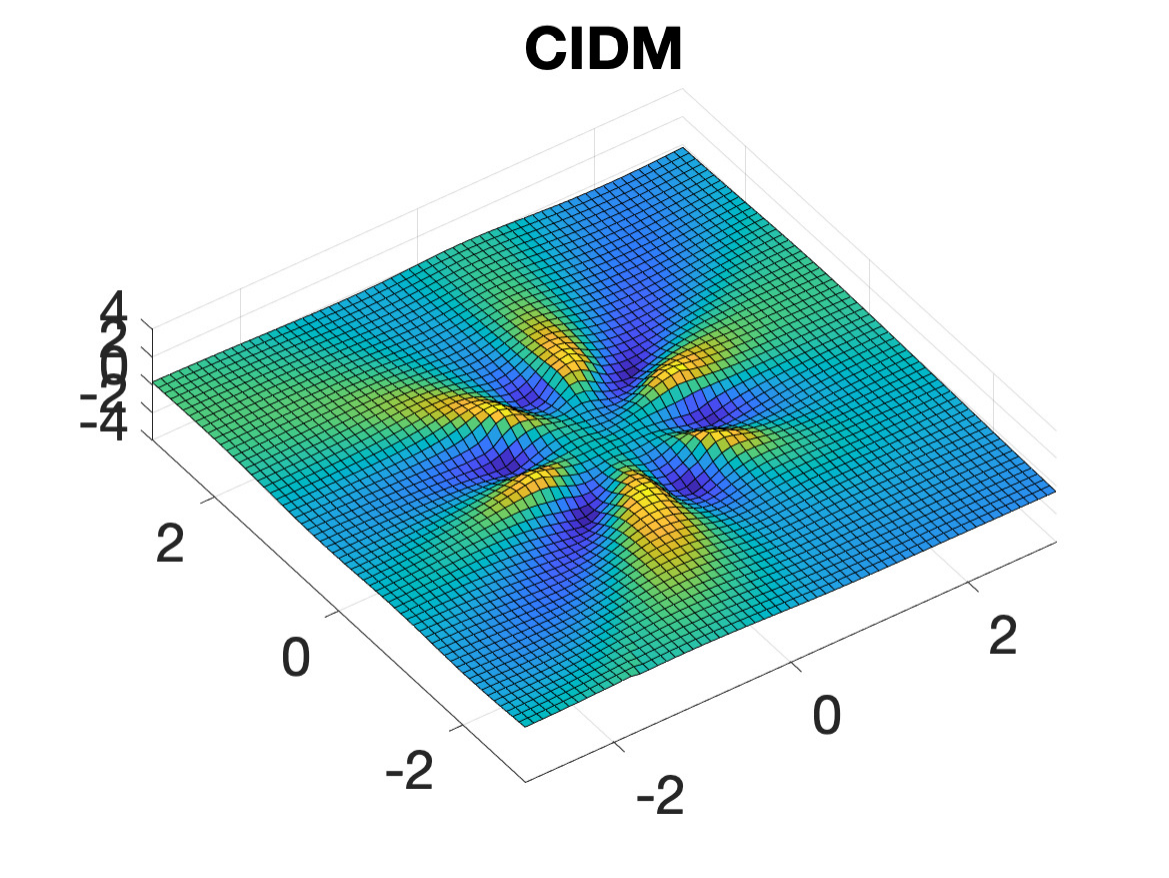}
    \includegraphics[width=0.24\linewidth]{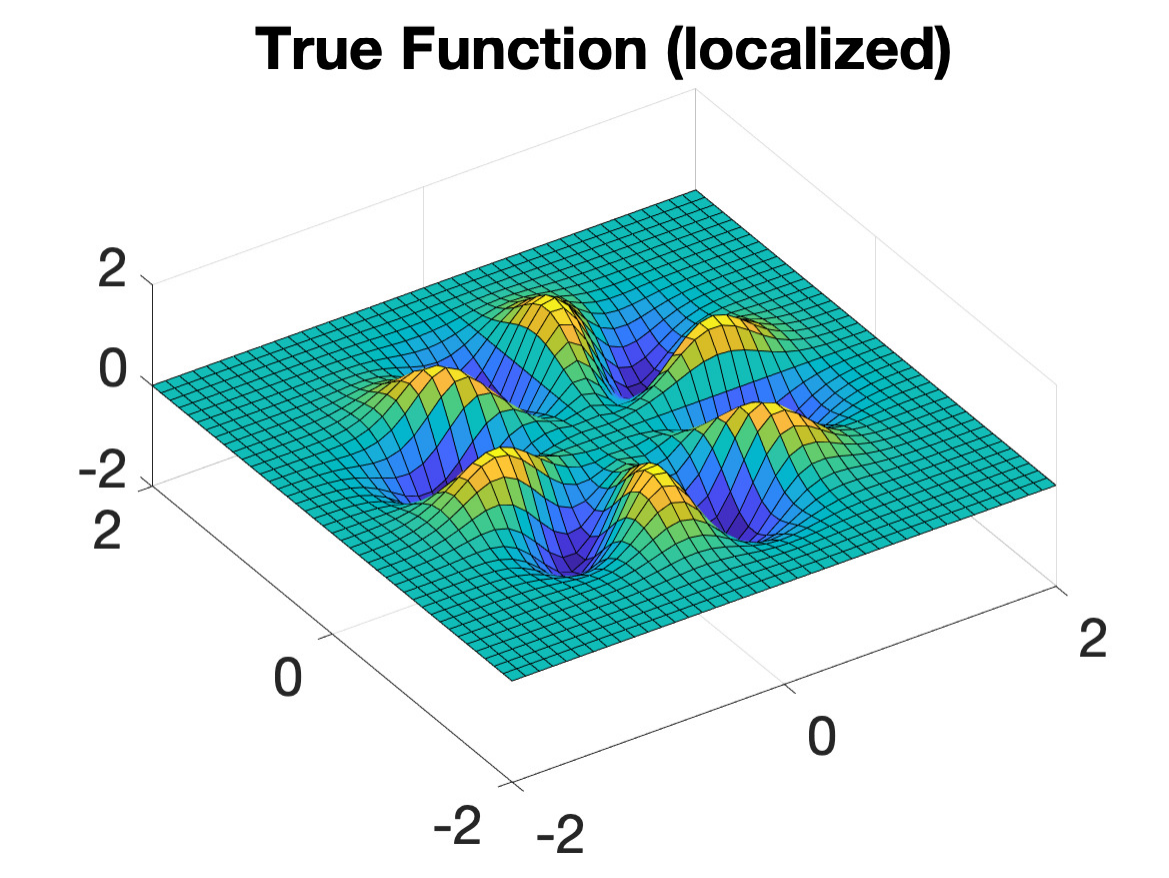}\includegraphics[width=0.24\linewidth]{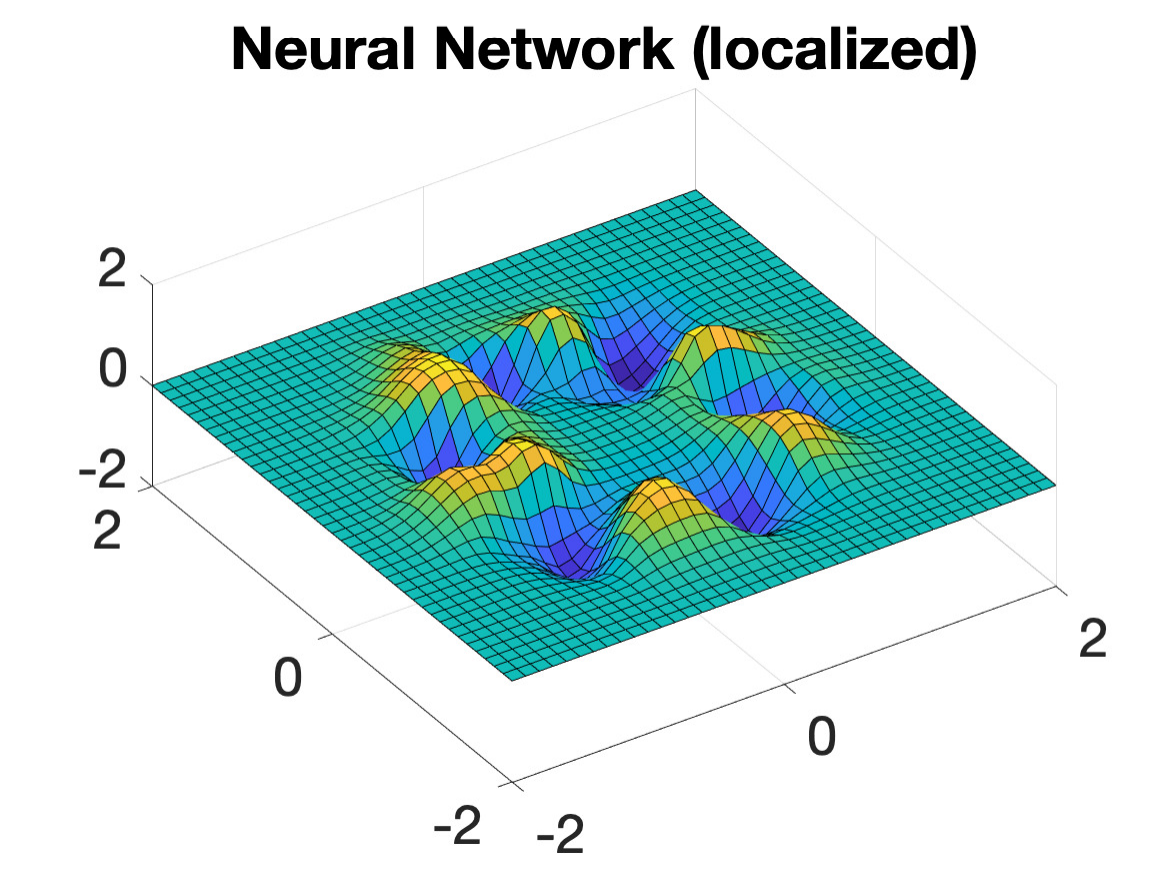}
    \includegraphics[width=0.24\linewidth]{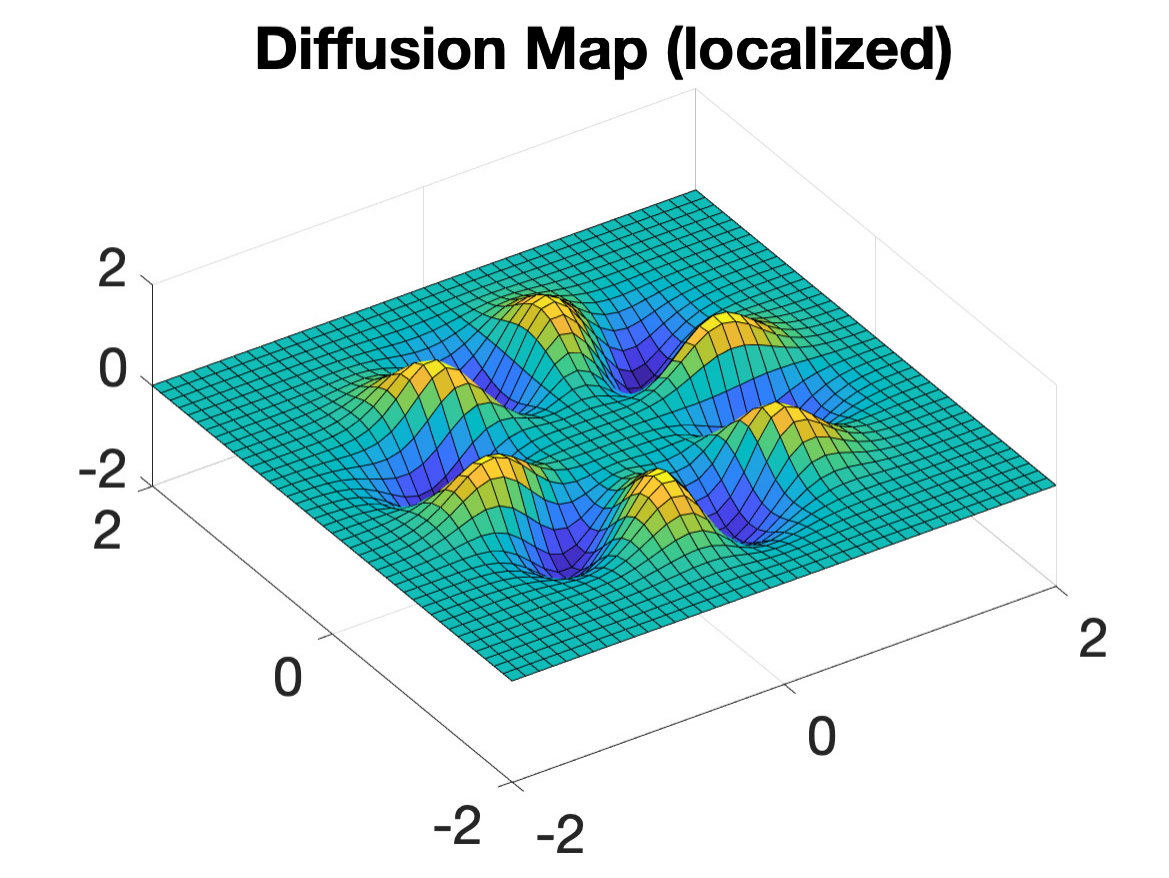}\includegraphics[width=0.24\linewidth]{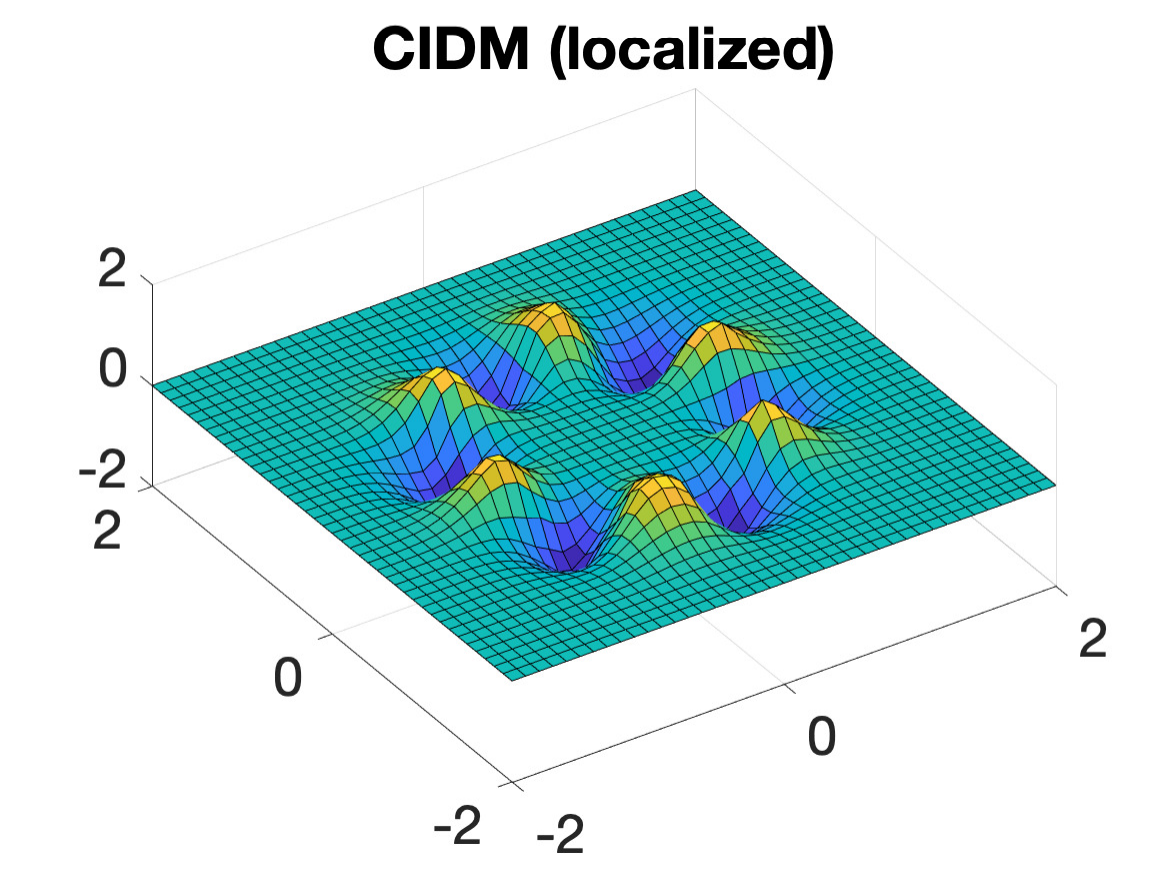}
    \caption{ \label{fig:Nystrom} \textbf{Nystr\"{o}m extension comparison}. Consider data points near a unit circle (top left) and a function to learn given by the color (also shown bottom left localized near the unit circle. We consider three methods of learning the function, a simple 2-layer neural net, the standard Diffusion Map Nystr\"{o}m extension, and the CIDM Nystr\"{o}m extension. Each extension is shown on a large region (top row) as well as localized near the unit circle (bottom row). The CIDM provides the smoothest extension to the entire input space.}
\end{figure}

A key property of the Nystr\"{o}m extension is that on the training data points we have,
\[ \phi(x_i) = \frac{1}{\lambda}\sum_{j=1}^N k(x_i,x_j)(\vec\phi)_j = \frac{1}{\lambda}(K\vec\phi)_i = \frac{1}{\lambda}(\lambda\vec\phi)_i = (\vec\phi)_i. \]
So if we interpret $(\vec \phi)_i$ as the value of a function on the data point $x_i$ then the Nystr\"{o}m extension agrees with these function values on the original data set and extends the function to the entire input space.  

Moreover, given an arbitrary vector of function values $\vec f_i$ on the data set, we can extend this function to the entire data set by representing $\vec f_i$ in the basis of eigenvectors and then applying the Nystr\"{o}m extension to these eigenvectors.  Let $\{ \phi_\ell \}_{\ell=1}^N$ be the collection of eigenvectors of the kernel matrix $K$.  Note that $(\vec\phi_\ell)_i$ will refer to the $i$-th entry of the $\ell$-th eigenvector.  Notice that,
\[ \vec f = \sum_{\ell=1}^N \left<\vec f, \vec\phi_\ell\right> \vec\phi_\ell \]
so if we replace the vector $\vec\phi_\ell$ with the Nystr\"{o}m extension we have the Nystr\"{o}m extension of $f$ given by,
\[ f(x) = \sum_{\ell=1}^N \left<\vec f, \vec\phi_\ell\right> \phi_\ell(x) = \sum_{\ell=1}^N \left<\vec f, \vec\phi_\ell\right> \frac{1}{\lambda_\ell}\sum_{j=1}^N k(x,x_j)(\vec\phi_\ell)_j.  \]
Notice that this can be viewed as a kernel extension of $f$ by rewriting the above as,
\[ f(x) = \sum_{j=1}^N k(x,x_j) \left( \sum_{\ell=1}^N \left<\vec f, \vec\phi_\ell\right> \frac{1}{\lambda_\ell} (\vec\phi_\ell)_j \right)\]
in other words the Nystr\"{o}m extension of a function is given by $f(x) = \sum_{j=1}^N k(x,x_j)c_j$  which is a linear combination of the kernel basis functions $k(\cdot,x_i)$, with coefficients $c_j \equiv  \sum_{\ell=1}^N \left<\vec f, \vec\phi_\ell\right> \frac{1}{\lambda_\ell} (\vec\phi_\ell)_j $.  This formula can be truncated for $\ell = 1,...,L$ with $L < N$ to get a smoothed, low-pass filtered version of the function.  When all of the eigenvectors are used, we have,
\[ f(x_i) = \sum_{\ell=1}^N \left<\vec f, \vec\phi_\ell\right> \frac{1}{\lambda_\ell}\sum_{j=1}^N k(x_i,x_j)(\vec\phi_\ell)_j  = \sum_{\ell=1}^N \left<\vec f, \vec\phi_\ell\right>(\vec\phi_\ell)_i  = \vec f_i \]
so again the Nystr\"{o}m extension agrees with the original vector of function values on the original data points. When fewer than $N$ eigenfunctions are used, the Nystr\"{o}m extension is a smoothing of the original function, so it does not interpolate the values on the training data, which can be useful for de-noising.  Finally, if we substitute in the definition of the vector inner product, we have the following expression for the Nystr\"{o}m extension,
\[ f(x) = \sum_{j=1}^N k(x,x_j) \left( \sum_{\ell=1}^L \frac{1}{\lambda_\ell}(\vec\phi_\ell)_j \sum_{i=1}^N\vec f_i (\vec\phi_\ell)_i   \right)\]
where $L$ is the number of eigenfunctions used and is typically much less than $N$ in order to smooth and denoise the function.

\section{Methods} \label{sec:methods}

Here we introduce some tools for analyzing data on manifolds in the input data space. The first tool is a novel method of projecting arbitrary data points nonlinearly down onto the manifold, the method is based on using the Nystr\"{o}m extension to build a projection, so we call this new method the \emph{Nystr\"{o}m Projection} in Section \ref{sec:nystrom_projection}.  The next tool is the Spectral Exterior Calculus (SEC) which was developed in \cite{berry_spectral_2020} and is able to identify vector fields that respect the global structure of the data. Here we overview the interpretation of the SEC on vector fields in Section \ref{sec:sec_vectors}, and describe how we use these vector fields to approximate the tangent space to the manifold in a way that is more robust than local linear methods.  Together, by using linear projection of vectors (such as perturbation directions) onto the tangent space, and the nonlinear Nystr\"{o}m Projection of perturbations of data points down onto the manifold itself, we introduce an on-manifold technique for Projected Gradient Descent in Section \ref{sec:on_manifold_pgd}.

\subsection{The Nystr\"{o}m Projection: Mapping Off-Manifold Points onto the Manifold}\label{sec:nystrom_projection}

The next (and crucial) question is: How does the Nystr\"{o}m extension perform out-of-sample?  In the case of manifold learning, this question has two cases, first, when the out-of-sample data lie on the manifold, and, second, when they are off the manifold (and potentially far from the manifold).  For data points on the manifold, the behavior of Nystr\"{o}m is well understood as a band-limited interpolation of the function $f$ which minimizes a certain cost function.  The on manifold out-of-sample interpretation is easy because we started by assuming that there was a given function on the manifold and that we had sampled values of that function on our in-sample training data.  Thus, there is a natural `true' function in the background to compare our Nystr\"{o}m interpolation to.

\begin{figure}[!h]
    \centering
    \includegraphics[width=0.3\linewidth]{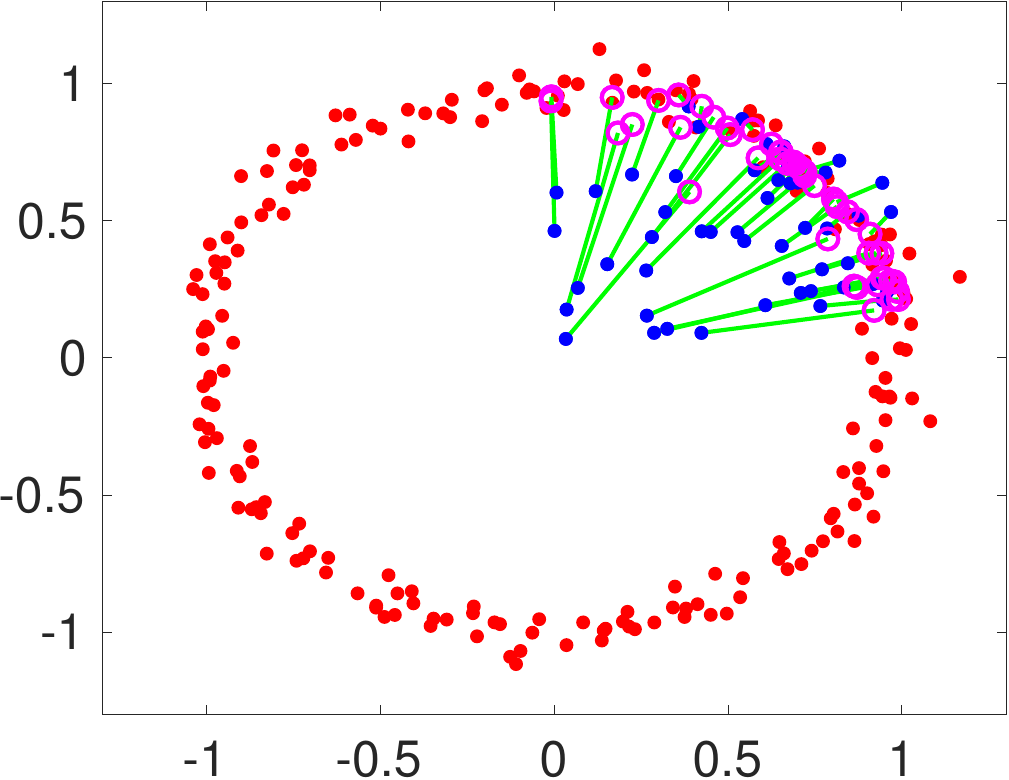}\includegraphics[width=0.3\linewidth]{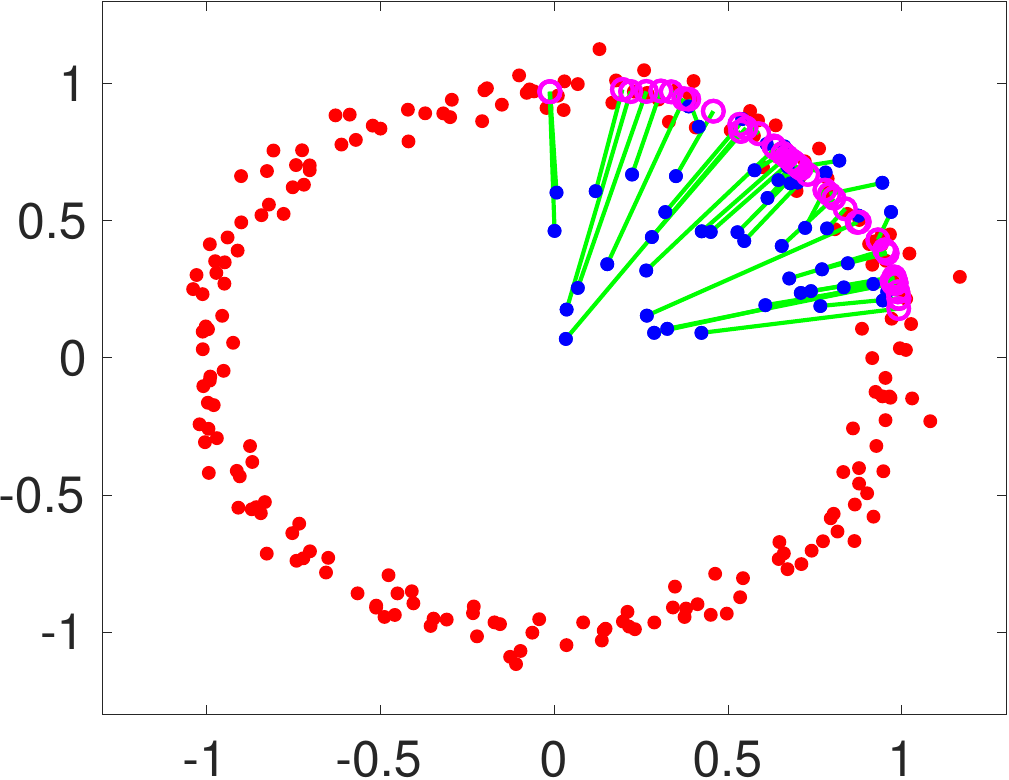}\includegraphics[width=0.3\linewidth]{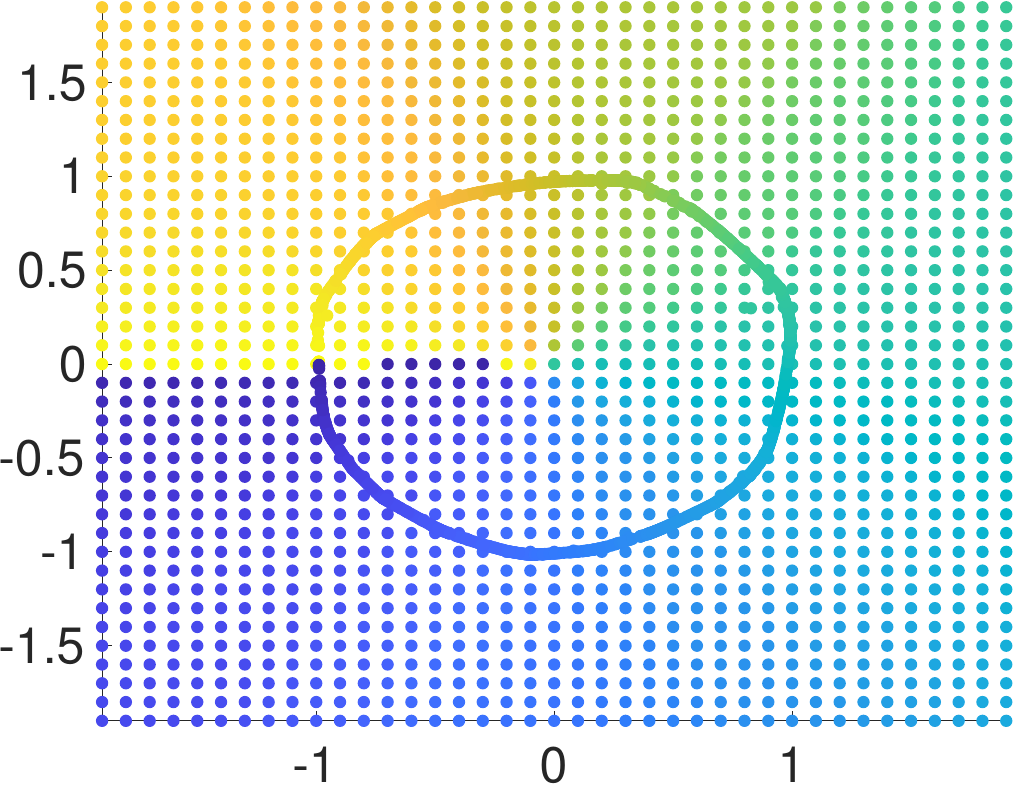}
    \includegraphics[width=0.3\linewidth]{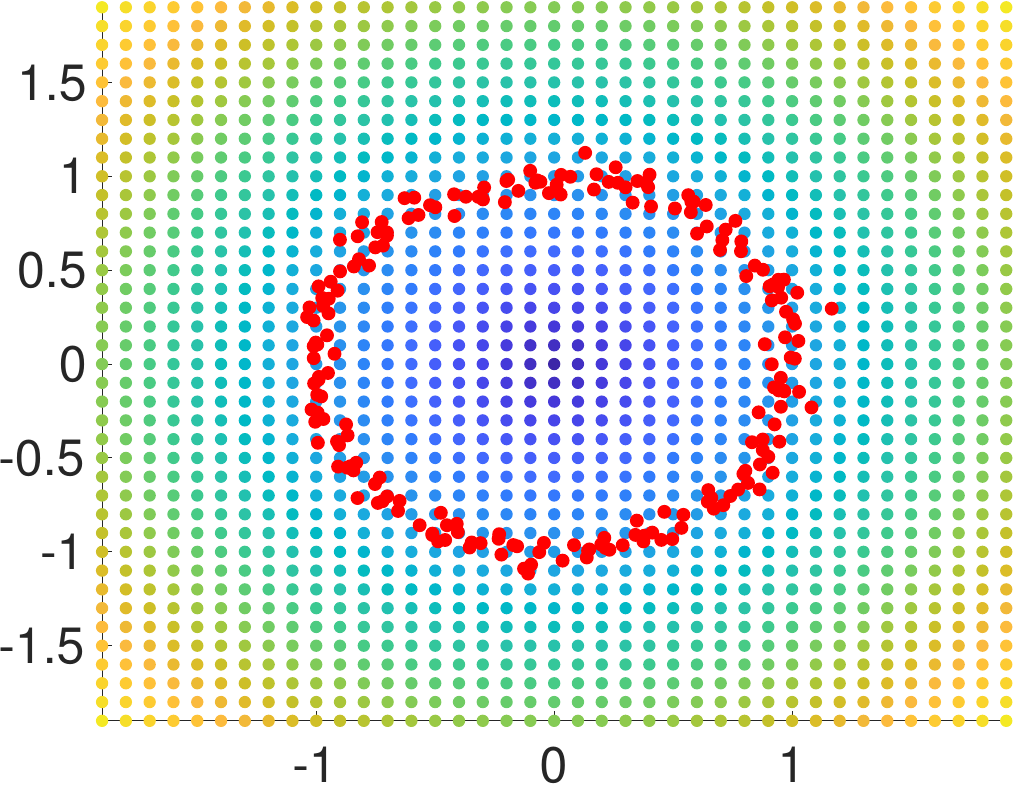}\includegraphics[width=0.3\linewidth]{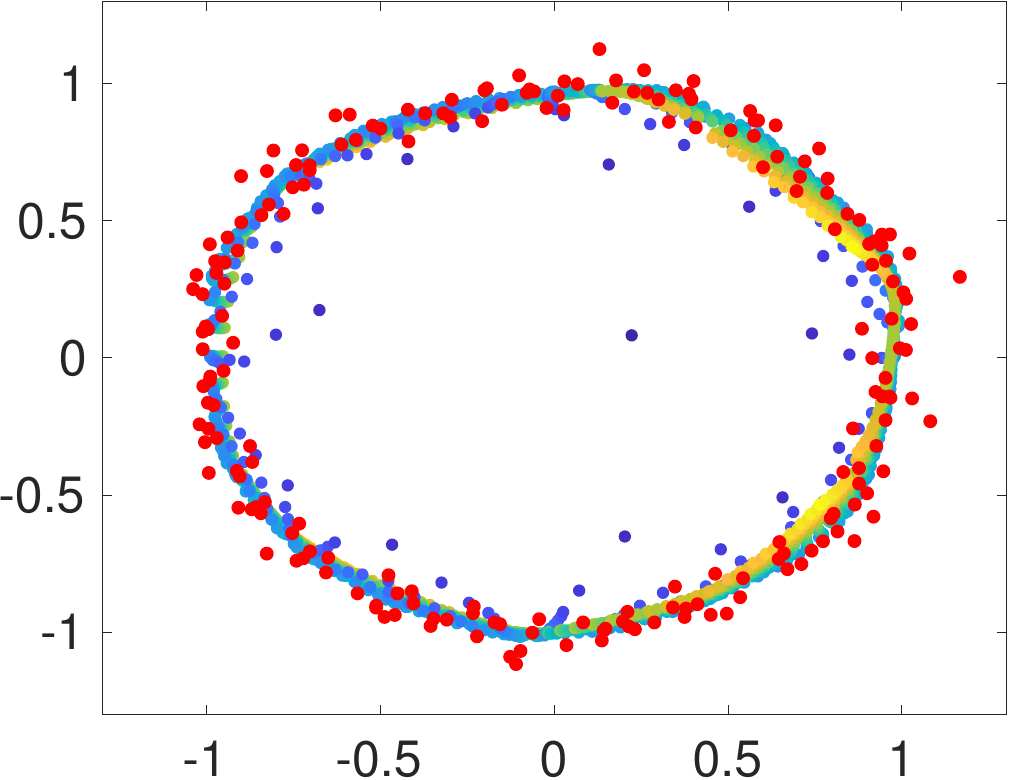}\includegraphics[width=0.3\linewidth]{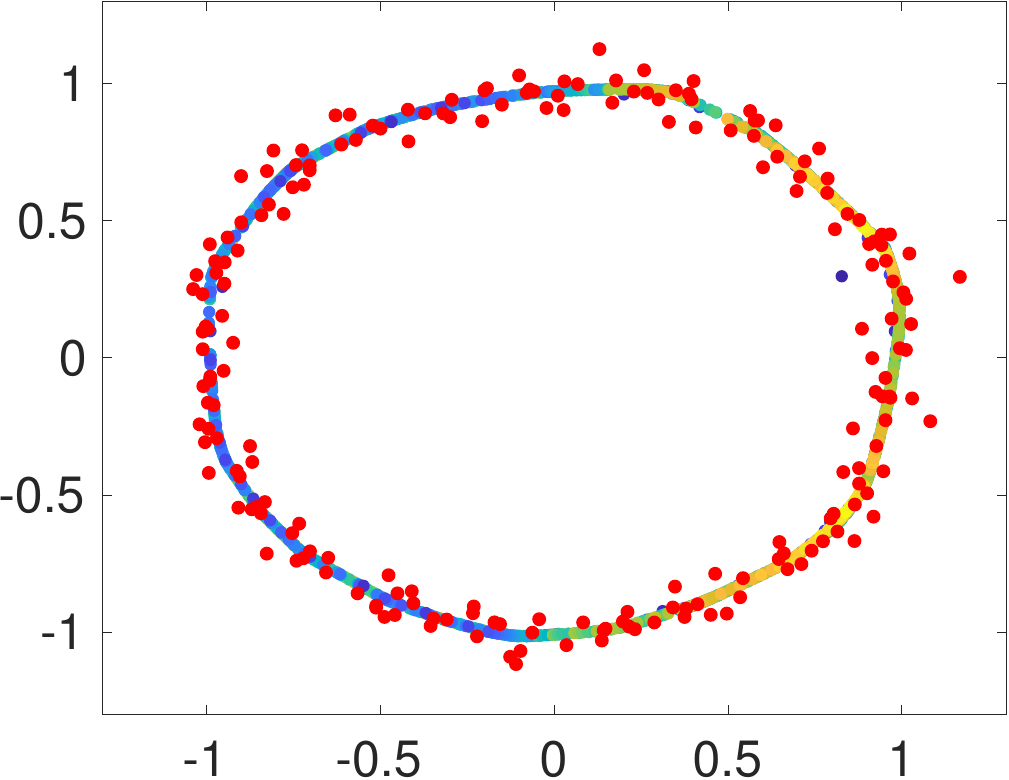}
    \caption{ \label{fig:nystromprojection} \textbf{Nystr\"{o}m Projection onto the Unit Circle}.  We use the Nystr\"{o}m Projection to project points on the plane onto the unit circle using a noisy training data set (red) to learn the manifold. Top: After learning from the red training data set, blue data points far from the manifold are projected onto the magenta points using a single iteration of the Nystr\"{o}m projection (left) and two iterations (middle), the green line connects each initial point to its Nystr\"{o}m Projection. Top Right: Applying two iterations to a grid in the plane projects onto a circle, the original grid points are colored by the angle computer after they are projected in order to show where they land.  Bottom Row: A grid colored by radius (left) is projected once (middle) and twice (right) using the Nystr\"{o}m Projection learned from the same data set (red) as the top row.}
\end{figure}

The case of extension to off-manifold points is much more interesting and less is known about this case.  Clearly, for any fixed `true' function defined on the manifold, there will be infinitely many smooth extensions to the entire space, so the Nystr\"{o}m extension is selecting one of these extensions, and in the limit of infinitely many data points and eigenfunctions, this extension is minimizing a certain functional.  While this is an open area of research, empirically we observe that for a normalized CIDM kernel, the Nystr\"{o}m extends the function to an off manifold data point by essentially taking the function value of the nearest point on the manifold.  Since almost every point in the ambient data space has a unique nearest point on the manifold, this is well defined up to a set of measure zero, and in practice there is a smoothing effect in a neighborhood of this measure zero set, however, we will ignore these effects for simplicity.  

To demonstrate empirically how the Nystr\"{o}m extension performs far from the training data set, in Fig.~\ref{fig:Nystrom} we show an example of a data set lying near the unit circle in the plane.  Given a simple smooth function, shown in the first panel of Fig.~\ref{fig:Nystrom}, we can use various methods to learn this function and attempt to extend it to the entire input data space (the plane in this case).  Notice that when well tuned the performance near the training data set, shown by the ``localized" panels of Fig.~\ref{fig:Nystrom} is comparable for a simple two-layer neural network as well as the Nystr\"{o}m extension with both the standard Diffusion Maps kernel and the CIDM kernel.  However, Fig.~\ref{fig:Nystrom} shows that these methods have very different behavior far from the data set, with the neural network behaving somewhat unpredictably, and the standard Diffusion Map kernel having difficulty extrapolating when far from the training data, whereas the CIDM makes a smooth choice of extension which is well-defined even very far from the training data.

This interpretation of the Nystr\"{o}m extension as taking the value of the nearest point on the manifold is critical since it lead us to a novel and powerful method of achieving a nonlinear projection onto the manifold.  The idea is actually quite simple, think of the original data set as a function on the manifold and build the Nystr\"{o}m extension of this function.  In fact, this is how we often think of a data set mathematically in the manifold learning literature.  Thus, we will apply the Nystr\"{o}m extension of the original data coordinates into a function on the entire data space, and we call the resulting function the \emph{Nystr\"{o}m Projection}.

While it is perfectly valid to consider the data manifold as a subset of the ambient data space, $\mathcal{M} \subset \mathbb{R}^n$ in differential geometry it is useful to think of an abstract manifold $\mathcal{N}$ that is simply an abstract set of points, and then think of the data set as the image of this abstract manifold under an embedding into Euclidean space, so $\iota:\mathcal{N} \to \mathcal{M}\subset\mathbb{R}^n$.  So now the points in data space, $x_i\in \mathbb{R}^n$, are each the images of an abstract point $\tilde x_i \in \mathcal{N}$ such that $\iota(\tilde x_i) = x_i$.  In this interpretation, each of the coordinates of the data are actually scalar valued component functions of the embedding function, so $(x_i)_s = \iota_s(\tilde x_i)$ where $\iota_s : \mathcal{N} \to \mathbb{R}$ are the component functions of the embedding.  Of course, since we know the value of these coordinate functions on the training data set, we can apply the Nystr\"{o}m extension to each of the $\iota_s$ functions, and extend the entire $\iota$ embedding map to the entire data space. In this way we obtain the Nystr\"{o}m Projection $\tilde\iota :\mathbb{R}^n \to \mathbb{R}^n$, which is given by,
\[ \tilde\iota_s(x) = \sum_{j=1}^N k(x,x_j) \left( \sum_{\ell=1}^L \frac{1}{\lambda_\ell}(\vec\phi_\ell)_j \sum_{i=1}^N (\vec x_i)_s (\vec\phi_\ell)_i   \right)\]
where $(\vec x_i)_s$ is the $s$-th coordinate of the $i$-th data point.  We can also write the Nystr\"{o}m Projection more compactly in terms of the Nystr\"{o}m extension of the eigenfunctions, as
\[ \tilde \iota(x) = \sum_{\ell=1}^L \hat x_\ell \phi_\ell(x)\]
where $\hat x_\ell = \left<\vec x, \vec \phi_\ell\right>$ is a vector-valued generalized Fourier coefficient given by $(\hat x_\ell)_s = \left<(\vec x)_s, \vec \phi_\ell\right> = \sum_{i=1}^N (\vec x_i)_s (\vec\phi_\ell)_i$.  Thus, we can think of $\hat x$ as encoding the embedding function that takes the abstract manifold to the realized data coordinates.  

In fact, $\tilde \iota$ does much more than the original embedding function, since it extends the embedding function to the entire data space.  This is because when input data points are off-manifold, the Nystr\"{o}m extension of a function is well approximated by selecting the values of the function for the nearest point on the manifold.  Since we are applying the Nystr\"{o}m extension to the embedding function itself, this means that for an off-manifold point, the Nystr\"{o}m Projection $\tilde \iota$ will actually return the coordinates of the nearest point on the manifold.  In other words, Nystr\"{o}m Projection $\tilde \iota$ acts as the identity for points on the manifold and projects off-manifold points down to the nearest point on manifold.  This novel yet simple tool gives us a powerful new ability in manifold learning and has opened up several promising new research directions.  Moreover, the choice of $L$ in the Nystr\"{o}m projection gives us control over the resolution of the manifold we wish to project on.  Thus, for a noisy data we can intentionally choose a smaller $L$ value in order to project down through the noise to a manifold that cuts through the noisy data.  This is demonstrated in Fig.~\ref{fig:nystromprojection}(rightmost panels) where we set $L=20$ and recovered a smooth circle that cuts through the noisy input data set (red points).

It is useful to formulate the Nystr\"{o}m Projection as a composition of a nonlinear map (related to the Diffusion Map) and a linear map.  Often we consider the map which takes an input point in data space and returns the coordinates of the first $L$ eigenfunctions, $\Phi:\mathbb{R}^n \to \mathbb{R}^L$ given by $\Phi(x) = (\phi_1(x),...,\phi_L(x))^\top$.  This is actually the so-called \emph{Diffusion Map} (with $t=0$). 
 While we have argued above that this is not necessarily the best embedding of the data, it is useful to express the projection we have just constructed.  Note that the $\hat x$ is an $n\times L$ matrix containing the first $L$ generalized Fourier coefficients of each of the $n$ coordinates of our data set. Thus, $\hat x$ defines a linear map $\hat X : \mathbb{R}^L \to \mathbb{R}^n$ from $\mathbb{R}^L$ (the image of $\Phi$) back to the data space, $\mathbb{R}^n$ (where $\hat X$ is simply given by left multiplication by the matrix $\hat x$).  The Nystr\"{o}m Projection onto the manifold is the composition of these two maps, namely, $\tilde \iota = \hat X \circ \Phi$ so that we have a map,
\[ x \mapsto_\Phi (\phi_1(x),...,\phi_L(x))^\top \mapsto_{\hat{X}} \tilde \iota(x) \]
such that $\tilde \iota \equiv \hat X \circ \Phi : \mathbb{R}^n \to \mathcal{M}\subset\mathbb{R}^n$ and $\left. \hat X \circ \Phi \right|_{\mathcal{M}} = {\rm Id}_{\mathcal{M}}$.  

Before continuing we consider a future application of the Nystr\"{o}m Projection as an input layer to a neural network. If we are performing optimization with respect to a loss function $L:\mathbb{R}^n \to \mathbb{R}$ and we want to restrict our optimization to the manifold, we can simply compose with the projection $\tilde \iota$ to find,
\[ L|_{\mathcal{M}}(x) = L(\hat X \circ \Phi)(x) \]
which has gradient,
\begin{equation}
\nabla L|_{\mathcal{M}}(x) = D\Phi(x)^\top \hat x^\top \nabla L((\hat X \circ \Phi)(x)) \in T_x\mathcal{M} \subset \mathbb{R}^n. 
\label{eq:grad_LM}
\end{equation}
Here we assume that the gradient of $L$ is already computable, and we are merely evaluating $\grad L$ on $\tilde\iota(x) = \hat X \circ \Phi(x)$ which is still a point in data space and just happens to have been projected down onto the manifold.  Moreover, we already have the matrix $\hat x$, so the only additional component that is needed is the gradient of $\Phi$, which simply requires computing the gradient of each of the Nystr\"{o}m eigenfunctions.

\subsection{SEC Vectors} \label{sec:sec_vectors}

Our goal is to find vector fields that span the tangents spaces of the manifold at each point.  While this is not always possible with just $d$ vector fields (where $d$ is the intrinsic dimension of the manifold), the Whitney embedding theorem guarantees that it is always possible with $2d$ vector fields.  The $L^2$ inner product, 
\[ G(v,w) \equiv \left<v,w\right>_{L^2} \equiv \int_\mathcal{M} v\cdot w \, d\textup{vol} \]
induced on vector fields by the Riemannian metric provides a natural notion of orthogonality that can help to identify non-redundant vector fields.  Here we use the notation $v\cdot w \equiv g(v,w)$ to denote the function which at each point is the Riemannian dot product of the two vectors at fields at that point.

\begin{figure}[!h]
    \centering
    \includegraphics[width=0.24\linewidth]{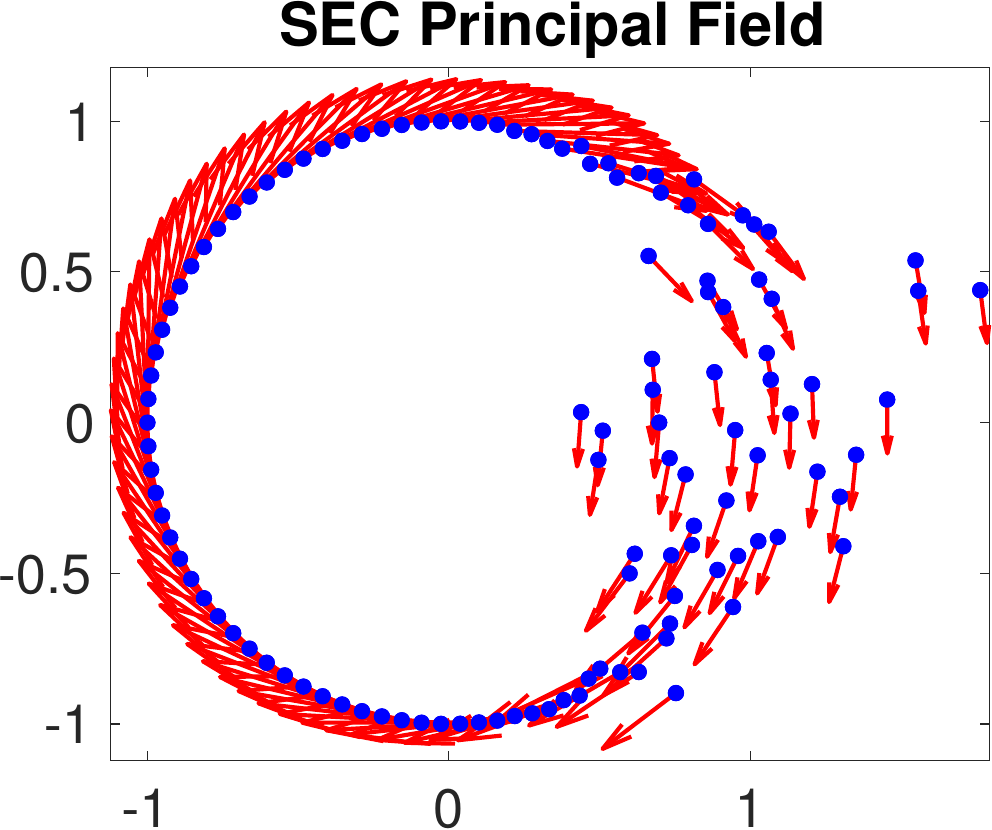}\includegraphics[width=0.24\linewidth]{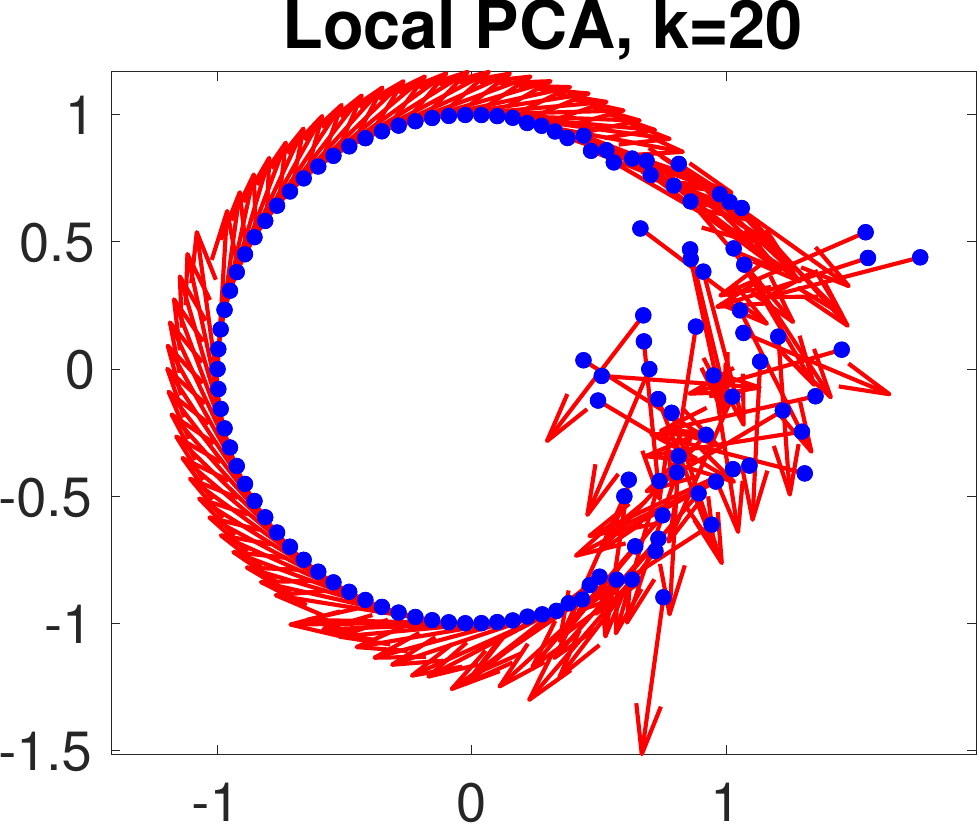}
    \includegraphics[width=0.24\linewidth]{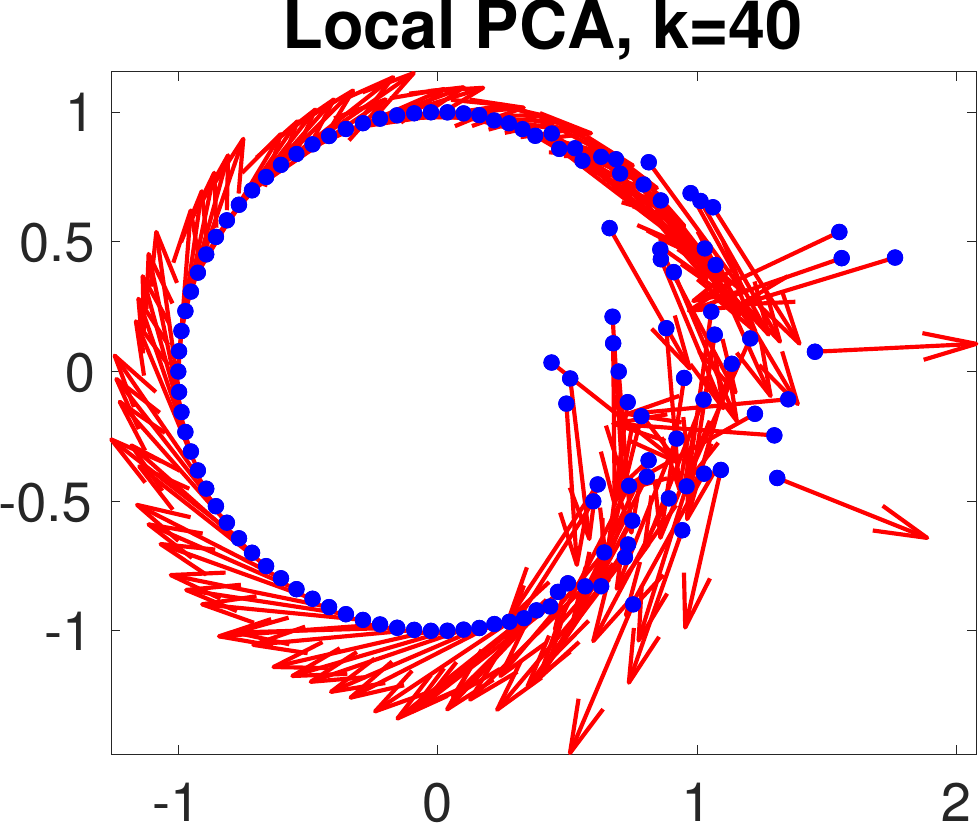}\includegraphics[width=0.24\linewidth]{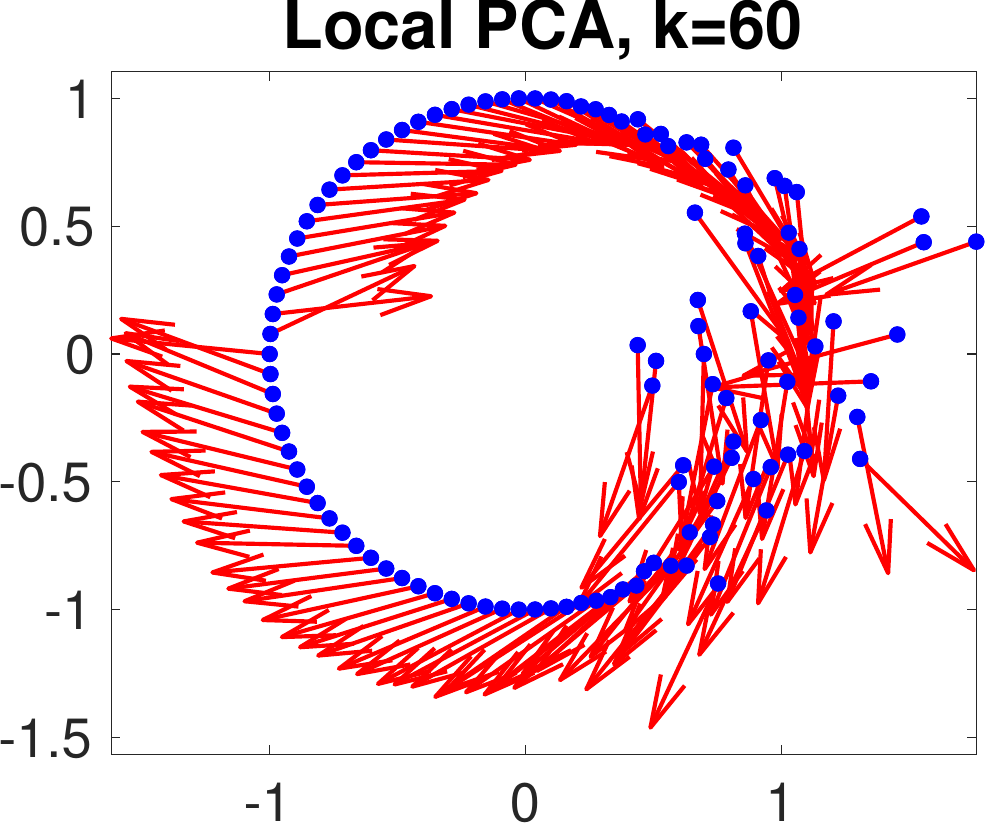}
    \includegraphics[width=0.24\linewidth]{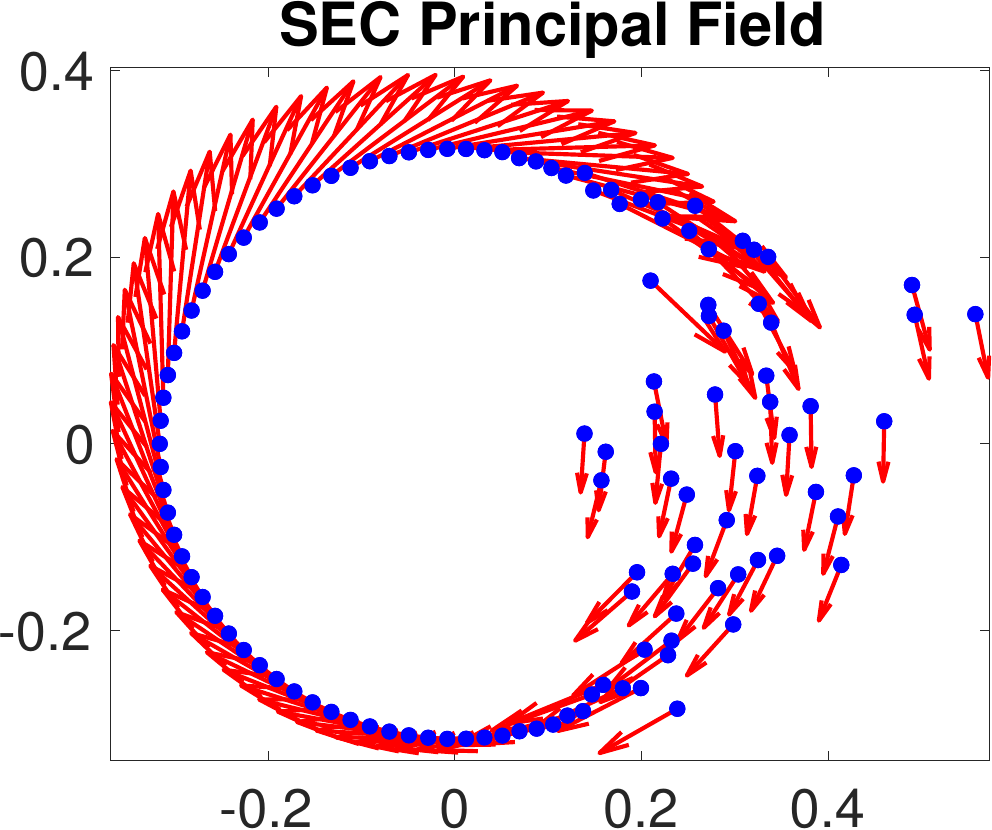}\includegraphics[width=0.24\linewidth]{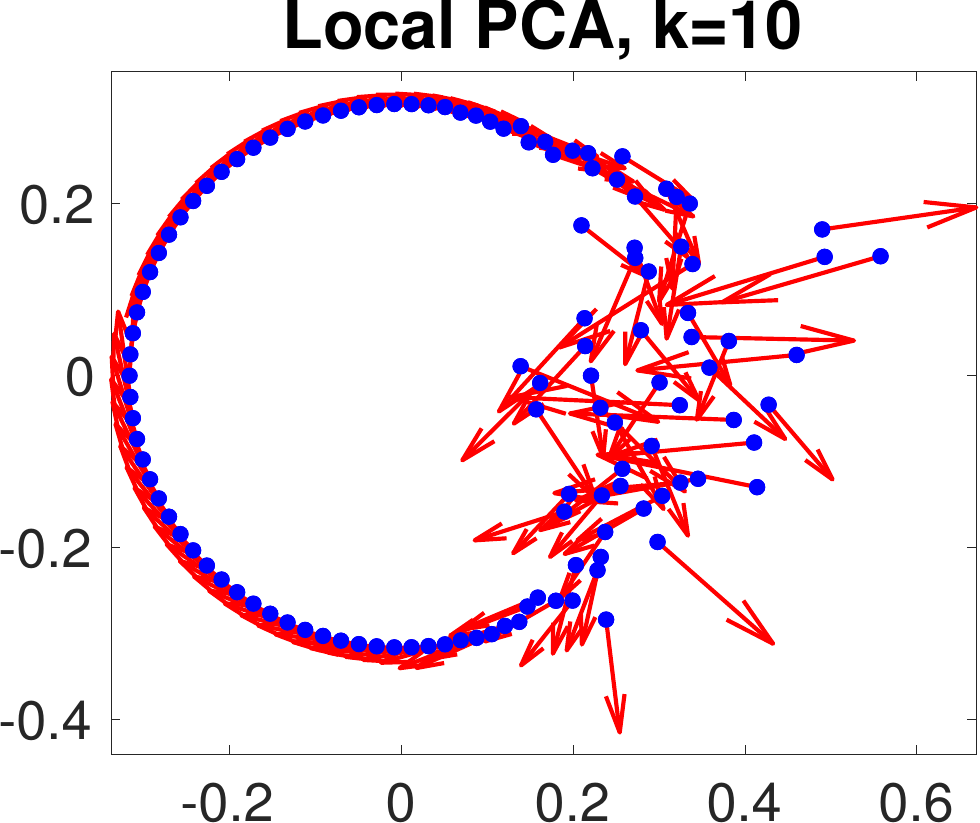}
    \includegraphics[width=0.24\linewidth]{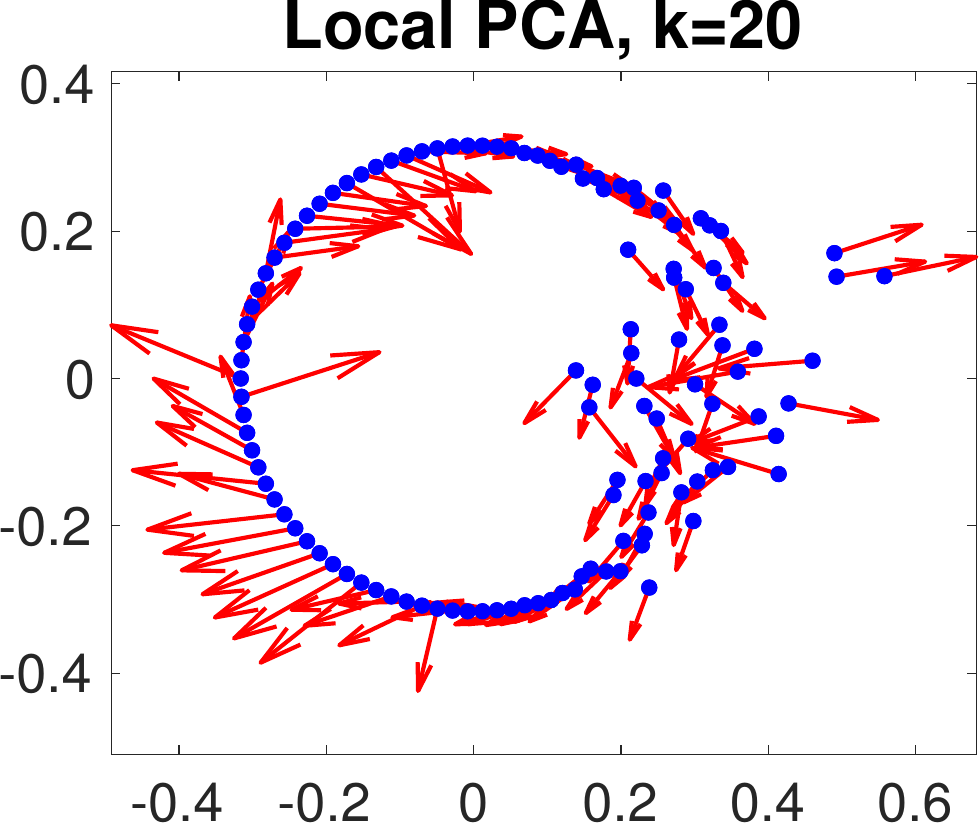}\includegraphics[width=0.24\linewidth]{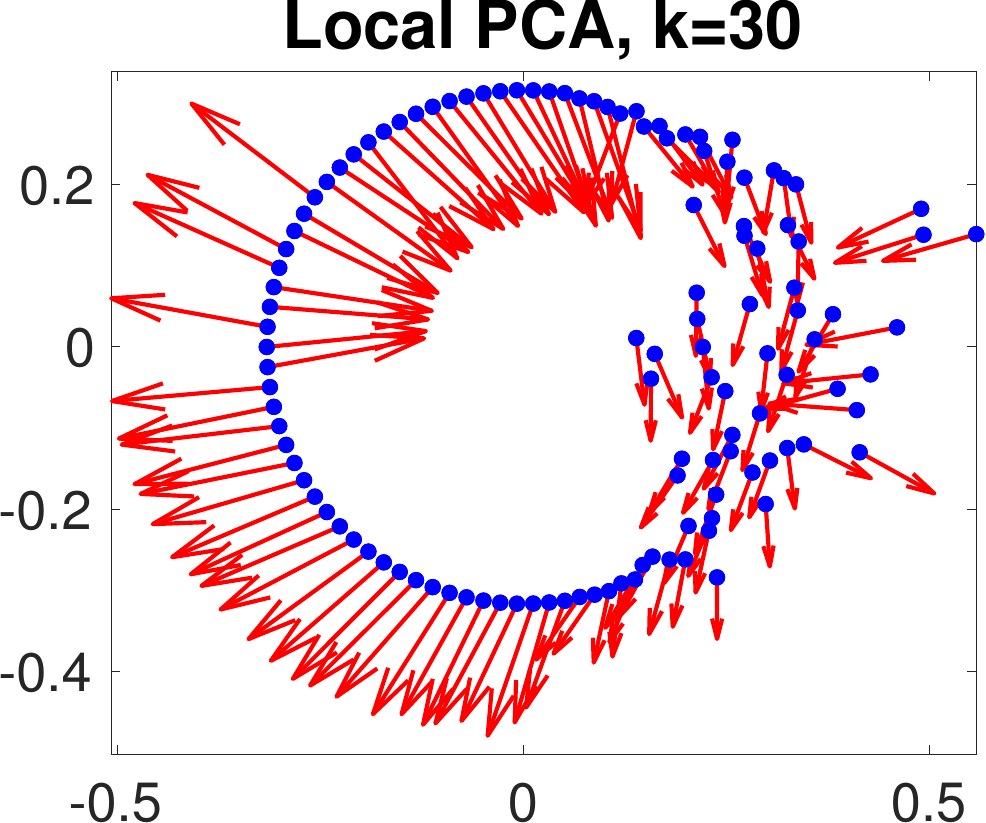}
    \caption{ \label{fig:SECvf} \textbf{SEC vector fields (left) respect global structure}. Here we consider a data set (blue points) localized near the unit circle but with varying density and varying amounts of noise.  In the top left plot we show the principal vector field identified by the SEC (the global minimizer of the Dirichlet energy). We compare this to a local PCA approach using k=20, 40, and 60 nearest neighbors of each point (middle left to right). Note how the SEC vector field smoothly respects the dominant mode of variation in the data set; while the local PCA approach can find the tangent direction in the clean sections, it loses track in the noisy section of the data.  The problem is further exacerbated in higher dimensions, in the bottom row we repeat the same experiment using an isometric embedding of the circle into four dimensions.  Again we show the SEC principal vector (bottom left) and local PCA with k=10, 20, and 30 nearest neighbors from middle left to right.}
\end{figure}

In addition to being orthogonal, we also need these vector fields to cut through noise and follow the coarse (principal) geometric structure of the manifold.  For this purpose, we introduce the Dirichlet energy on vector fields, induced by the weak form of the Hodge 1-Laplacian, $\delta_1 = d\delta + \delta d$ where $d,\delta$ are the exterior derivative and codifferential respectively.  These operators are defined on differential forms, which are dual to tensor fields, and in particular the dual of a vector field is a 1-form.  The musical isomorphisms switch back and forth between forms and fields, with sharp, $\sharp$, turning forms into fields, and flat, $\flat$, going back. As an example, the codifferential on 1-forms is related to the divergence operator by,
\[ \nabla \cdot v = -\delta(v^\flat) \]
Similarly, the exterior derivative, which acts on forms, induces an operator on smooth fields which generalizes the curl operator,
\[ \nabla \times v \equiv (\star d(v^\flat))^\sharp. \]
This operator coincides with the curl when manifold is 3-dimensional so we use the same name, but in general on an $n$-dimensional manifold the output of the generalized curl operator will be a $n-2$ tensor field.  Using the generalized divergence and curl operators, we can now define the Dirichlet energy on smooth vector fields,
\[ E(v,w) \equiv \int_\mathcal{M} (\nabla\times v) \cdot (\nabla\times w) \, d\textup{vol} + \int_\mathcal{M} (\nabla \cdot v)(\nabla \cdot w) \, d\textup{vol}  \]
where we note that the dot product in the first term is the extension of the Riemannian metric to $n-2$ forms.  By minimizing this energy, we will ensure that we have the smoothest possible vector fields, as seen by a global (integrated) measure of smoothness.  As discussed in Appendix \ref{sec}, the Dirichlet energy on vector fields is motivated by a dual energy on differential 1-forms.  Note that while measuring smoothness on functions only requires a single term, the integral of the gradient of the function, measuring smoothness of vector fields requires two different types of derivatives.  This is because neither the divergence nor the curl can completely measure the different types of oscillations a vector field can have, but when combined they provide a robust measure of smoothness.

The Dirichlet energy and Riemannian metric together will define a natural functional for identifying good sets of vector fields, which in turn will reduce to a generalized eigenvalue problem.  In Appendix \ref{sec} we overview the SEC construction of the Dirichlet energy and how to find its minimizers relative to the Riemannian inner product.  For now we demonstrate the advantage of this approach to finding vectors fields that respect the global structure of the data set.  In Fig.~\ref{fig:SECvf} we show a data set that, while near a simple manifold, exhibits variations in density and noise levels characteristic of real data. This example clearly demonstrates how the SEC principal vectors respect the principal global structure of the manifold, rather than getting lost in local details the way local linearization is.  This is possible because the Dirichlet energy captures a measure of smoothness that is balanced over the entire global structure of the data set.

%In the previous section we derived the global Riemannian metric tensor, $G$, in the SEC frame to enforce orthogonality (integrated).  We now overview the weak-form of the 1-Laplacian, to enforce smoothness.  
Finally we should note an important caveat and related direction for future research.  The vector fields identified here are only globally orthonormal, meaning in an integrated sense.  This means that they are not required to be orthonormal in each coordinate chart, since positive alignment in some regions can be cancelled by negative alignment in others.  In future work one could consider a local orthonormality condition.  This is related to the search for minimal embeddings.

\begin{figure}[!h]
    \centering
    \includegraphics[trim={0 0 0 0},clip,width=0.7\linewidth]{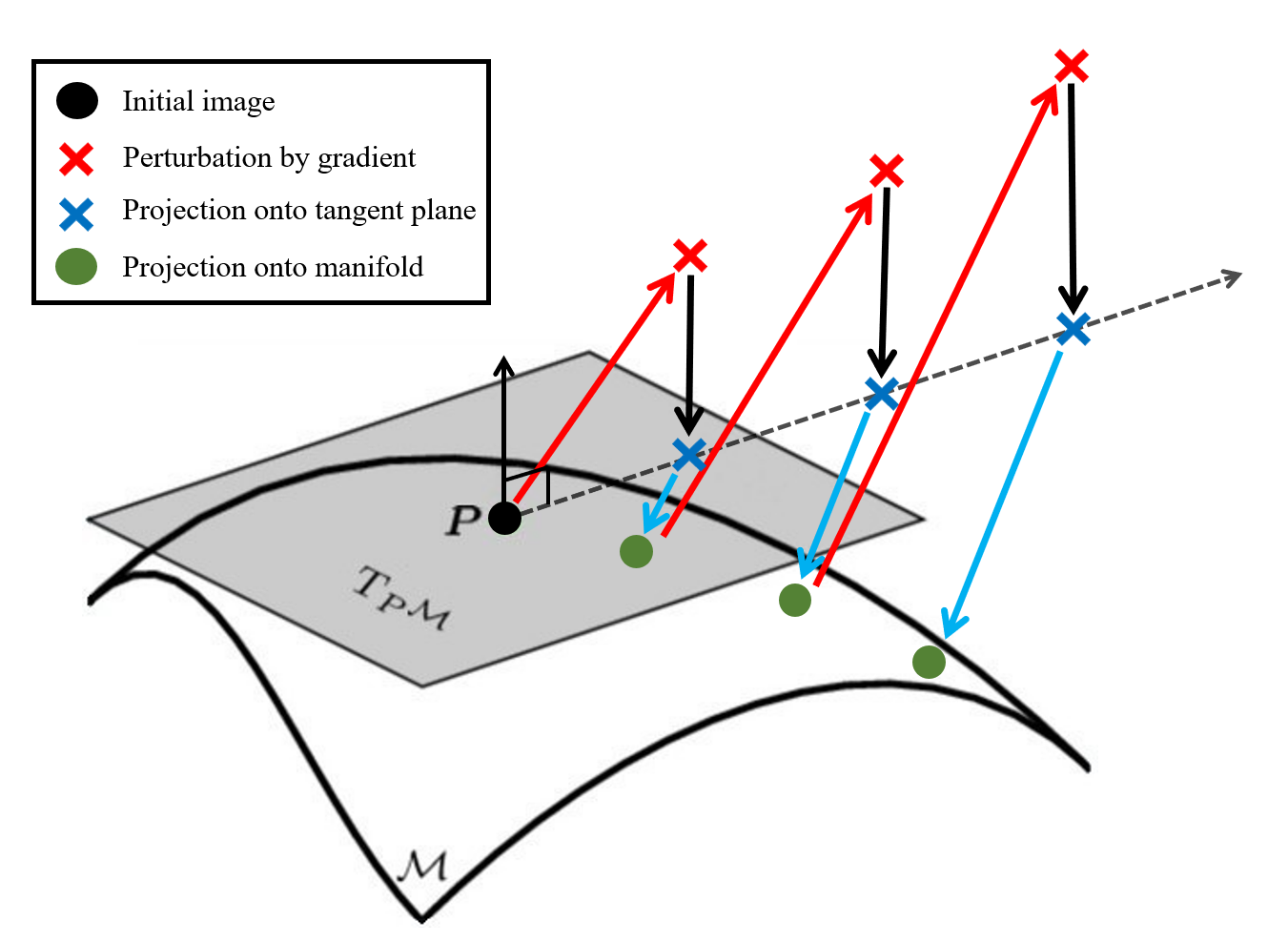}
    \caption{ \label{fig:om_pgd_steps} 
    \textbf{On-manifold PGD steps for a 2D tangent space}. Starting with an initial image (black dot), we follow the network gradient (red arrow). The perturbed sample (red X) is then projected (black arrow) onto the tangent plane, $T_P\mathcal{M}$ (gray surface), from the geometry found around the initial image. 
    The upwards black arrow pointing out of the tangent plane at the black dot is normal to the tangent plane for visual clarity.
    The on-plane perturbation (blue X) is then Nystr\"{o}m projected (blue arrow) onto the manifold (curved surface $\mathcal{M}$). The on-manifold point (green dot) is then classified to determine if the network has classified it correctly. If it is classified correctly then the process is repeated, starting with the on-manifold point from the previous step.
    }
\end{figure}

\subsection{On-Manifold Projected Gradient Descent} \label{sec:on_manifold_pgd}
% Standard PGD papers: \cite{tsipras_robustness_2019, santurkar_image_2019}
As discussed in \S\ref{sec:introduction}, Projected Gradient Descent (PGD) relies on computing the network loss gradient with respect to the input rather than the weights of the model. 
Once these input gradients have been computed via backpropagation with respect to a target class, the input is perturbed in the direction of the gradient to create an adversarial example. 
If the resulting perturbed image is farther than $\epsilon$ away from the starting image for a given distance metric (typically $\ell_2$ or $\ell_\infty$) then the perturbed image is projected onto the $\epsilon$-ball surface around the starting image.
This is done with the intent of keeping the adversarial example from becoming unrecognizable as the original class.
We use the SEC and Nystr\"{o}m projection to create an on-manifold PGD that is representative of the original class but not constrained to an $\epsilon$-ball.
First we take the gradient of the network for an input with respect to the input's class.
Then we find the input point's position on the manifold with the Nystr\"{o}m projection, see \S\ref{sec:nystrom_projection}.
Using the SEC we then compute the vectorfields that are tangent to the manifold and obtain the tangent bundle at the position of the starting point's position on the manifold, see Fig. \ref{fig:SECvf} for a comparison of vectorfields from SEC to local PCA.
Next we project the gradient onto the orthonormalized subspace spanned by a subset of tangent vectors at the input point depending on the dimensionality of the underlying data. 
For instance, if the underlying manifold had an estimated dimension of 2 then you would choose the 2 tangent vectors from the smoothest vectorfields computed by the SEC.
We orthonormalize the subspace by using the non-zero eigenvectors of the singulvar value decomposition (SVD) in order to obtain an unbiased basis.
Then we step along the direction of the gradient in the tangent space by an amount determined by a hyperparameter so that the input is sufficiently perturbed.
As the final step we use the Nystr\"{o}m projection to return this perturbed sample back to the manifold, see Fig \ref{fig:nystromprojection} for an example.
The Nystr\"{o}m projection ensures that the resulting example is on that particular class's manifold and provides the information for determining the example's semantic labels (intrinsic coordinates) and tangent vectors. 
We can obtain the on-manifold example's semantic labels using the same methodology used obtain the mapping from CIDM coordinates to pixel, see Fig. \ref{fig:Nystrom} for an example of obtaining semantic labels on a learned manifold. 
These steps are repeated until a misclassification is found, see Fig. \ref{fig:om_pgd_steps} for an generalized overview. 
If the on-manifold PGD is successfull then it produces an adversarial example that fools the classifier and is also on the input class's manifold.

\begin{figure}[!h]
    \centering
    \includegraphics[trim={0 0 0 0},clip,width=0.9\linewidth]{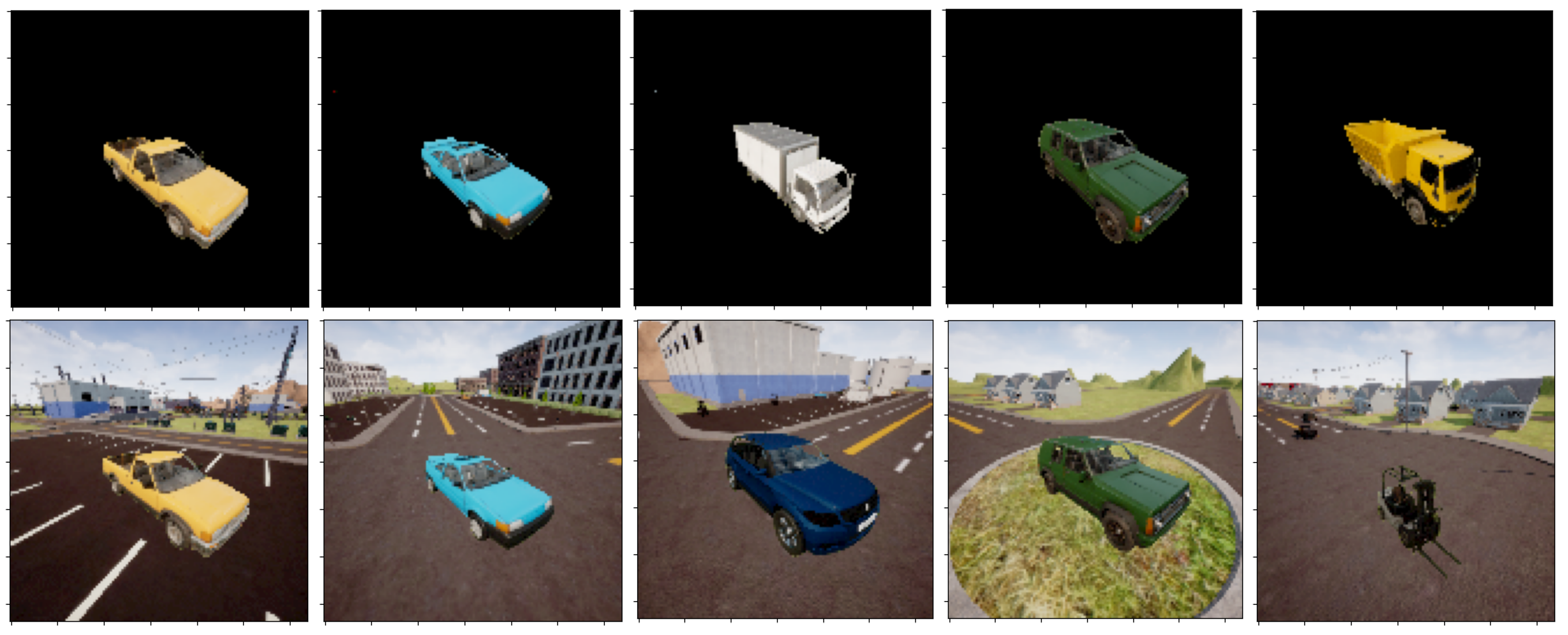}
    \caption{ \label{fig:rgb_data} 
    \textbf{Examples of RGB vehicle datatset}, consisting of 7 vehicle types (pickup truck, SUV, sedan, dump truck, box truck, jeep, fork lift), varying in 360 degrees azimuth and 45 degrees down look angle. The data contains 6 types of scene backgrounds including urban and rural environments. The full resolution data is 256 by 256 pixels, although it has been downsampled to 128 by 128. Top row: vehicle images with flat backgrounds generated using image segmentation maps. Bottom row: vehicle images with randomly sampled location backgrounds (images are again inserted into backgrounds using segmentation maps). }
\end{figure}

\section{Results} \label{sec:results}
\begin{figure}[!hb]
    \centering
    \includegraphics[trim={0 0 0 2cm},clip,width=0.9\linewidth]{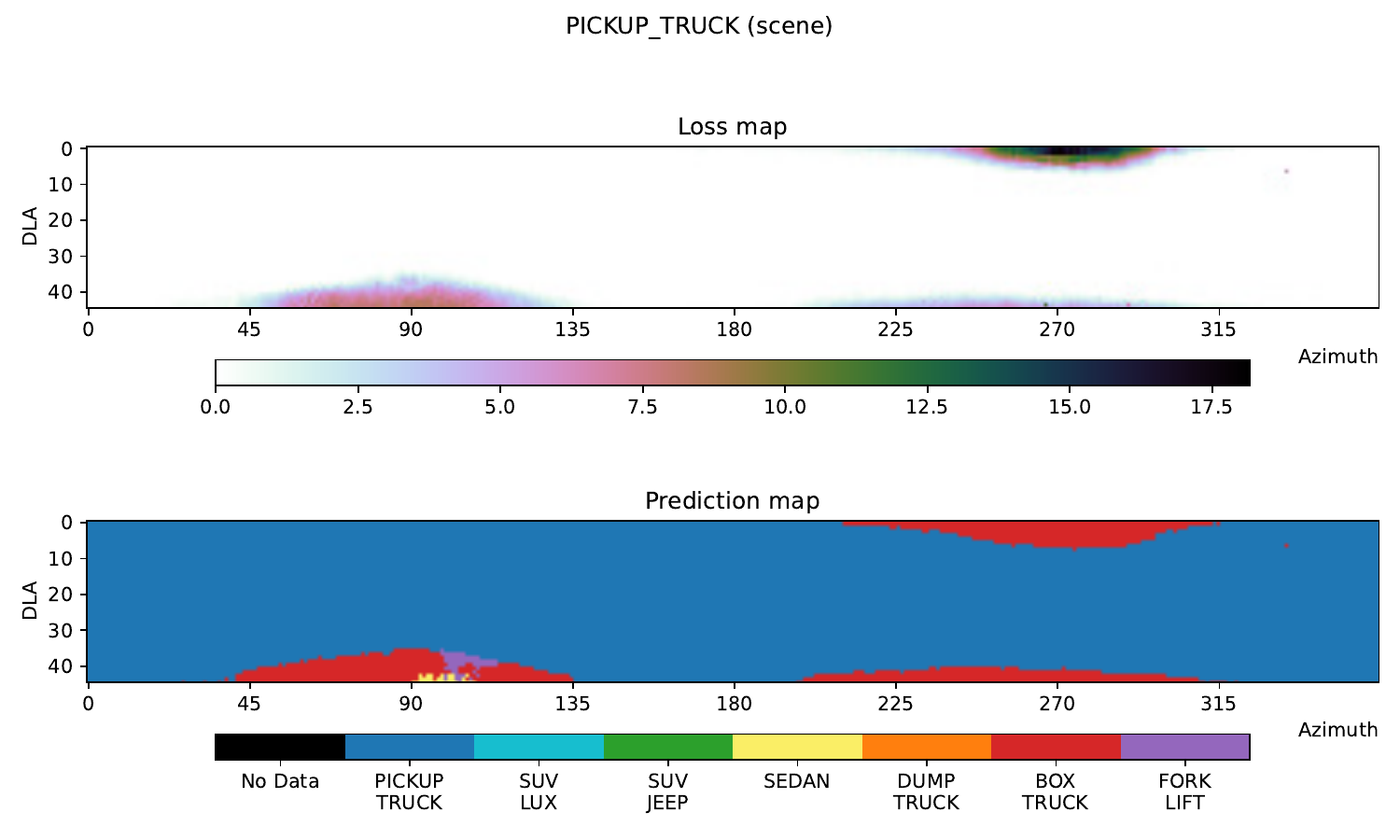}
    \caption{  \label{fig:vgg_loss_pred_map} 
    \textbf{Network loss and prediction map (pickup truck class). } Loss map and decision boundary map for the Pickup Truck class on a VGG11 network. The classifier was trained with the 7 vehicle types shown in Fig \ref{fig:rgb_data} using down look angles from $10^\circ - 30^\circ$ and azimuths from $0^\circ - 359^\circ$.   
    }
\end{figure}
% 1. Discuss network classification problem
% 2. Discuss vehicle image dataset 
% 3. Results for on manifold adversaries
% 4. Results for PGD and on manifold PGD
The main result we present shows on-manifold adversarial examples that are explainable in human understandable terms by using the adversaries' semantic labels on the manifold.
We present experimental results for finding on-manifold adversaries using a VGG11 classifier~\cite{simonyan2014very} and a synthetic dataset. 
Our classifier is trained and validated to classify RGB images of various vehicles (see Figure~\ref{fig:rgb_data} for examples), which were generated using a synthetic data collect in Microsoft's AirSim platform, allowing for insertion of various vehicle classes in a range of locations. 
Each vehicle class is sampled over two sets of intrinsic parameters, represented by the azimuth angle and down look angle (DLA) from which the image is captured. 
The azimuth angles are sampled one degree apart from 1 to 360 and the down look angle is sampled one degree apart from 1 to 45 degrees. 
The dense sampling in intrinsinc parameters will enable CIDM and SEC computations, while also allowing for NN models trained on the data to be fairly robust. 

\begin{figure}[!h]
    \centering
    \includegraphics[trim={0 0 0 1.6cm},clip,width=0.9\linewidth]{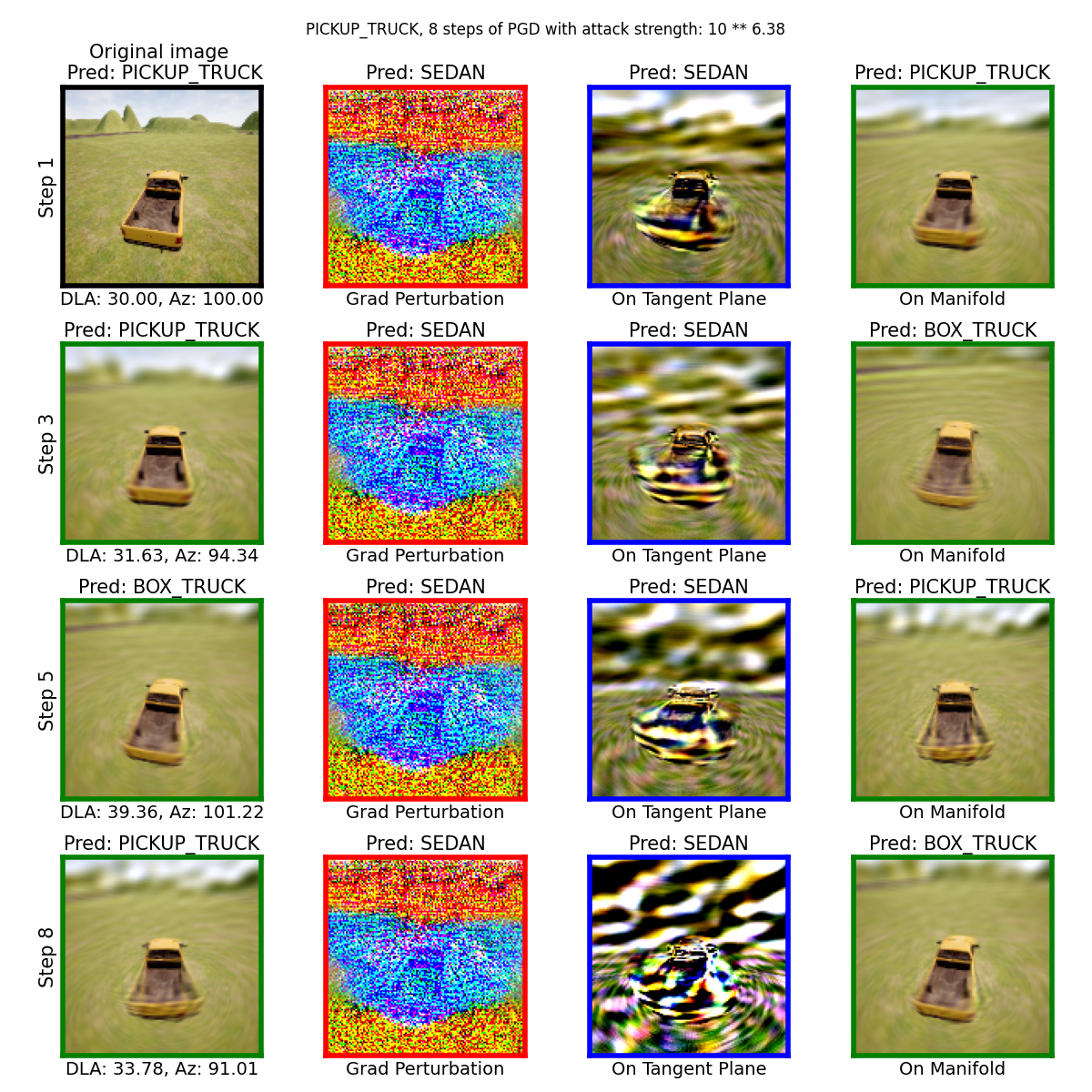}
    \caption{ \label{fig:PICKUP_on_man_adv_ex} 
    \textbf{Generation of an on-manifold adversarial image. } The first column is the Nyst\"{o}m projection of the input from the previous step with the step number on the vertical axis and the predicted intrinsic parameters on the horizontal axis. The second column is the first column after adding the network gradient times $10^{6.38}$. The third column adds the image from the first column and $10^7$ times the gradient after it is projected onto the manifold subspace spanned by the tangent subspace. The fourth column is the result of Nyst\"{o}m projecting the third column so that it is on the manifold. Each image includes the network prediction of that image at the top and the outline color corresponds to the steps in Fig. \ref{fig:om_pgd_steps}.
    }
\end{figure}

We trained a VGG11 classifier on a subset of the down look angles ($10^\circ - 30^\circ$) and all azimuth angles. 
See Fig. \ref{fig:vgg_loss_pred_map} for the loss and decision boundaries of the classifier for a single class over all view angles available, including those the classifier was not trained on.
We use an image from the training set that was in a region close to a decision boundary at $30^\circ$ down look and $100^\circ$ azimuth. 
We then apply our on-manifold PGD to that point using a manifold modeled on the points surrounding that point from $0^\circ - 40^\circ$ DLA and $80^\circ - 120^\circ$ azimuth, see Fig. \ref{fig:PICKUP_on_man_adv_ex}. 
The output of the on-manifold PGD for this example results in a misclassification that correlates with the decision boundary for this class near that sample in view angle as seen in Fig. \ref{fig:vgg_loss_pred_map}.
Note that the results of projecting onto the manifold provides not only the on-manifold point in pixel space, but also in intrinsic parameter coordinates.
This means that the misclassification can be explained in terms of human-understandable features.
In this case the on-manifold adversarial example seen at step 5 in Fig. \ref{fig:PICKUP_on_man_adv_ex} is shown to be at 39.36 DLA and 101.22 azimuth, which can be confirmed to be a region of misclassification by looking at the explicit sampling of that region in the decision boundary map of Fig. \ref{fig:vgg_loss_pred_map}.

\begin{figure}[!h]
    \centering
    \includegraphics[trim={0 0 0 1.1cm},clip,width=0.9\linewidth]{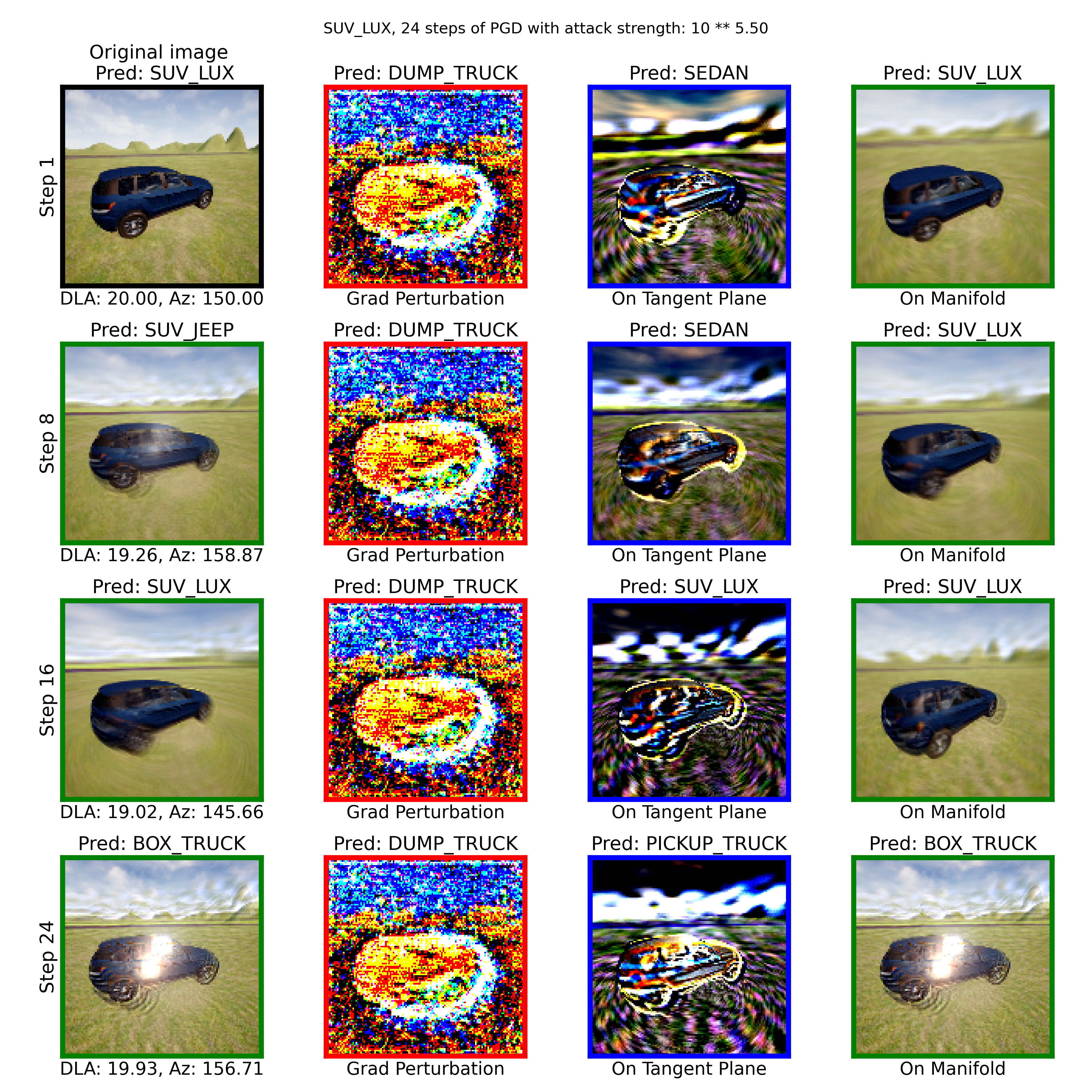}
    \caption{ \label{fig:SUV_LUX_on_man_adv_ex} 
    \textbf{Generation of an on-manifold adversarial image. } The first column is the Nyst\"{o}m projection of the input from the previous step with the step number on the vertical axis and the predicted intrinsic parameters on the horizontal axis. The second column is the first column after adding the network gradient times $10^{5.23}$. The third column shows the gradient after it is projected onto the manifold subspace spanned by the tangent subspace. The fourth column is the result of Nyst\"{o}m projecting the third column so that it is on the manifold. Each image includes the network prediction of that image at the top and the outline color corresponds to the steps in Fig. \ref{fig:om_pgd_steps}.
    }
\end{figure}

When perturbing the on manifold images with the gradient, $X' = X + \alpha \nabla X$, we use a fixed step size, $\alpha = 10^{6.38}$ for convenience. 
We tested a range of step sizes ranging on a log scale in order to find the smallest step that sufficiently perturbed the input into misclassifying with the on-manifold PGD algorithm. 
The size of the parameter $\alpha$ is due to the fact that the network gradient is mostly contained in a space not spanned by the tangent vectors.
After projecting the gradient onto the tangent plane, the magnitude of the projected gradient is several orders of magnitude smaller than the raw gradient,
which is visibly discernible in the second and third columns of Fig. \ref{fig:PICKUP_on_man_adv_ex} and \ref{fig:SUV_LUX_on_man_adv_ex}.
In order to obtain a sufficient perturbation for the classifier to misclass this required a large value for $\alpha$.
Due to the fact that the $\ell_2$ norm of the gradient was typically on the order of $10^{-3}$ for these examples, the gradient could also be normalized so that smaller values of $\alpha$ could be used.

\begin{figure}[!h]
    \centering
    \includegraphics[trim={0 0 0 0},clip,width=0.75\linewidth]{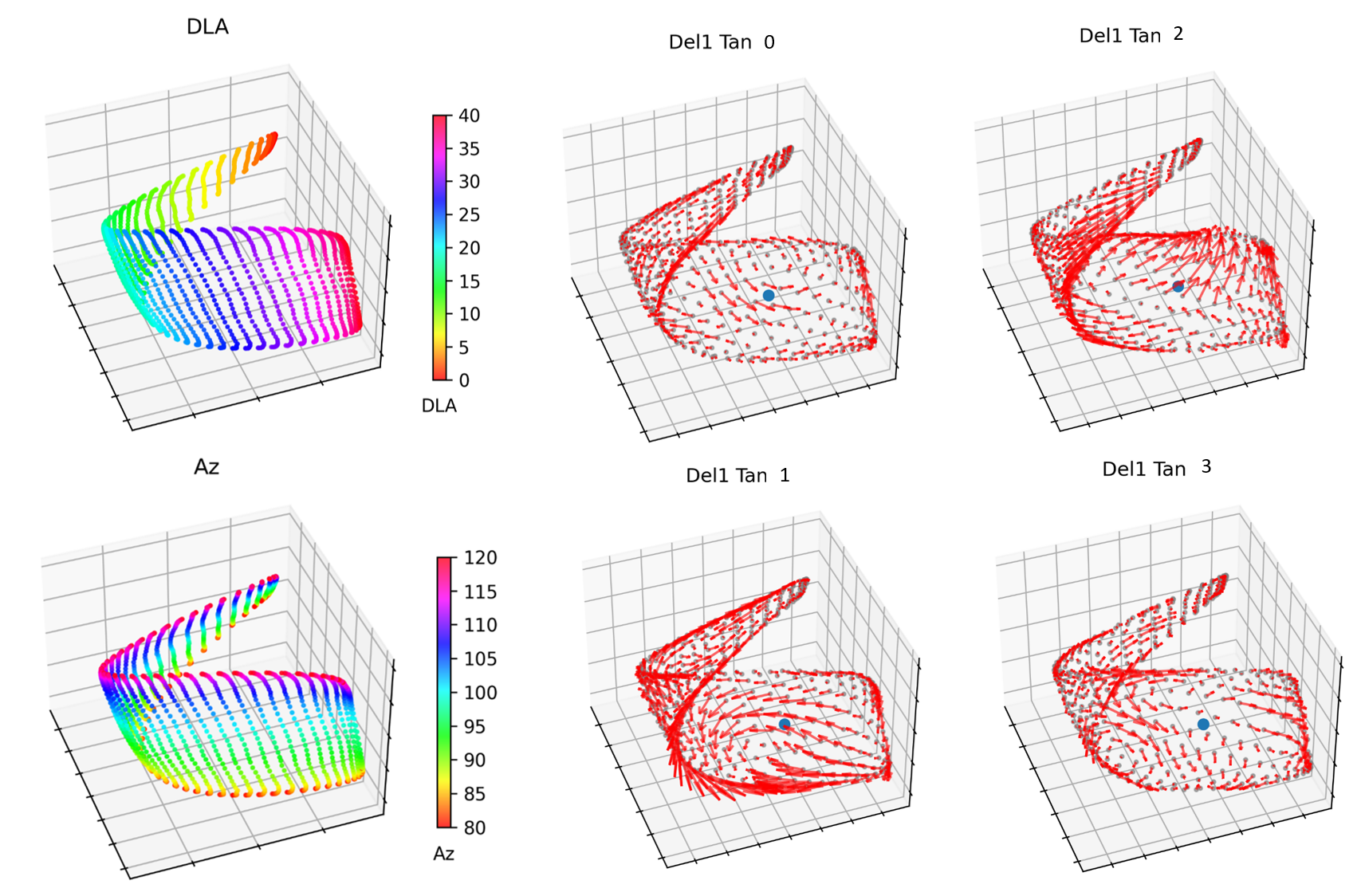}
    \caption{ \label{fig:truck_manifold} 
    \textbf{Pickup Truck manifold approximation.} 
    The top two images show the manifold of the Pickup Truck plotted using the first three coordinates from CIDM. 
    The top left image is colored by the azimuth angle, and the top right is colored by the down look angle.
    The bottom 4 plots show the tangent vectorfields of the SEC as red arrows, and the initial point around which the geometry approximation was built as a blue dot.
    }
\end{figure}

We choose the first two tangent vectors from the SEC because the output of the SEC code returns the vectorfields ordered by smoothness, which typically results in the first vectorfields being best aligned with the manifold. 
We visually confirmed this in CIDM coordinate space as seen in Fig. \ref{fig:truck_manifold}.

\begin{figure}[!h]
    \centering
    \includegraphics[trim={0 0 0 2cm},clip,width=0.9\linewidth]{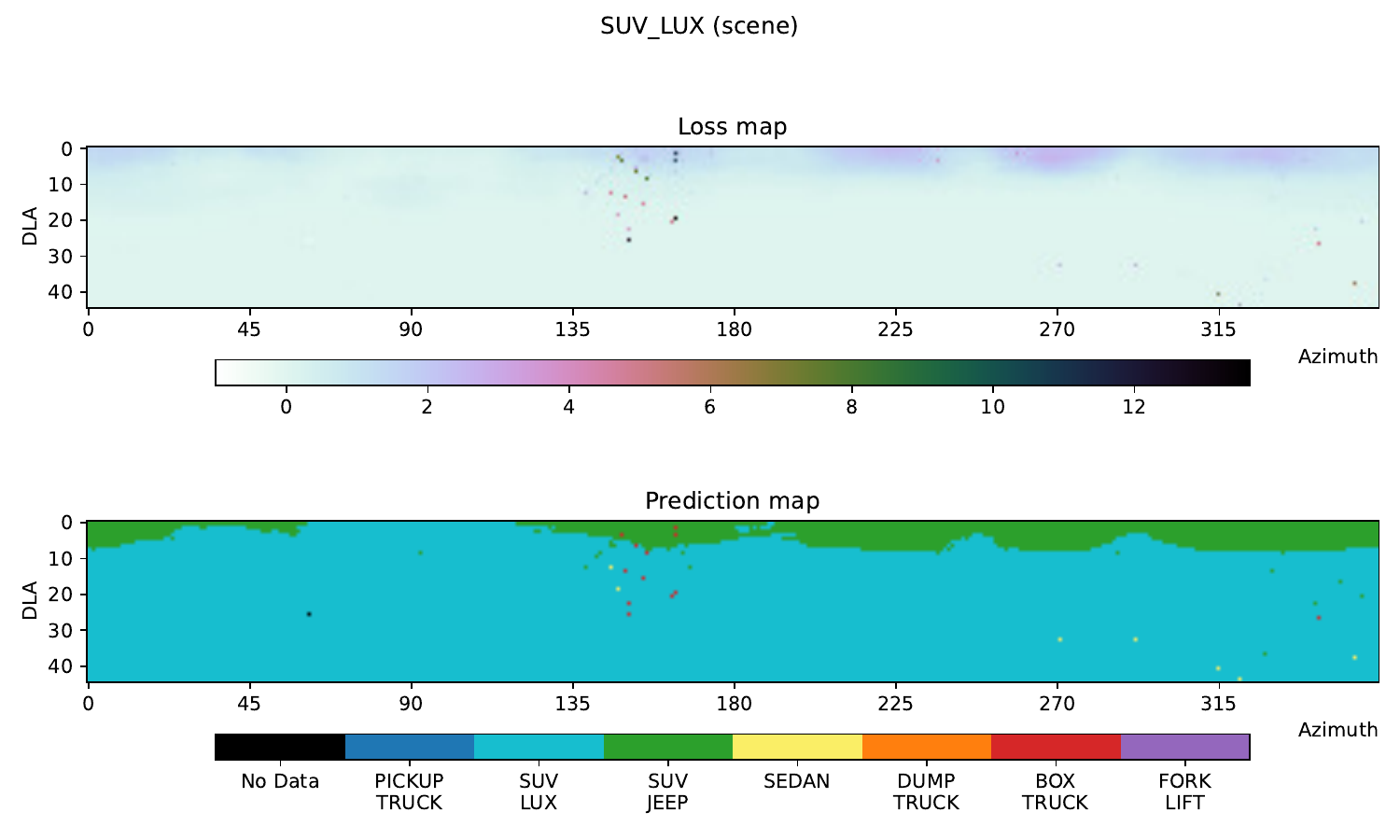}
    \caption{  \label{fig:cnn_loss_pred_map} 
    \textbf{Network loss and prediction map (luxury SUV class).  } Loss map and decision boundary map for the Luxury SUV class on a CNN. The classifier was trained with the 7 vehicle types shown in Fig \ref{fig:rgb_data} using down look angles from $20^\circ - 30^\circ$ and azimuths from $0^\circ - 359^\circ$.  Adversarial training begun after epoch 6 and continued until epoch 100, where the network gradients were used to generate adversaries. Network evaluation takes place with scene background chips, while the network was trained with a randomized set of backgrounds to avoid an over-dependence of the network on the image background.
    }
\end{figure}

We present another on-manifold PGD result using an adversarially trained network. The training paradigm for this experiment consists of using chips with randomly generated backgrounds, as shown in Fig. \ref{fig:rgb_data}, with the result that the trained network will be background-agnostic. We train on chips from all azimuth angles and DLA from 20 to 30 degrees. After 6 epochs of standard training, we switch to adversarial training, where adversarial chips are generated by following the network gradient. We continue until we reach 100 total epochs of training. Figure~\ref{fig:cnn_loss_pred_map} then plots the loss and prediction maps over azimuth and DLA for the luxury SUV class. 

We note in Figure~\ref{fig:cnn_loss_pred_map} that the majority of the misclassifications occur outside of the training regime, with a few rare misclassifications in the training regime, given as scattered predictions of  the box truck and sedan classes. We seek to generate on-manifold adversarial examples in the range $20^\circ - 30^\circ$ for DLA and $135^\circ - 165^\circ$, which is the range we use in geometry calculations. Notice, in this experiment, we are only providing our geometry tools with data from the intrinsic parameter range that the network was trained on. The goal of this experiment is to use our geometry tools to find and correctly explain an adversarial example that caused the network to mislcassify, without giving the geometry tools preferential treatment in the form of additional data that the network was not trained on. Figure~\ref{fig:SUV_LUX_on_man_adv_ex} illustrates our on-manifold PGD iteration, as described in Section~\ref{sec:on_manifold_pgd} with the starting point at DLA $20^\circ$ and azimuth $150^\circ$. 

Figure~\ref{fig:SUV_LUX_on_man_adv_ex} depicts the iteration of on-manifold PGD to an on-manifold adversarial example, which is classified as a box truck (as one would expect from Figure~\ref{fig:cnn_loss_pred_map}). In this experiment, we note that adversarial examples causing a misclassification often contain a reflection on the side of the vehicle. We have verified that this matches with a rare feature in the training data, where images containing a reflection commonly cause the luxury SUV to be misclassified as the white box truck. These reflections are a result of AirSim's environment and they represent an unexpected challenge that network mislcassified, but was identified by our on-manifold PGD approach. Evidently, not only is this approach able to generate adversarial example which can be explained in terms of their semantic labels, but it also provides explainable insights into which features of an image cause a network to misclassify.

\section{Discussion} \label{sec:discussion}
In conclusion, we have presented a formal introduction of CIDM, a type of variable bandwidth kernel diffusion maps that is adept at dealing with heterogenous data density. 
We have also introduced a novel application of the Nystr\"{o}m method for extending the CIDM eigenfunctions to new data points. 
We use the Nystr\"{o}m projection to map off-manifold points onto the manifold inside PGD to implement an on-manifold PGD.
Additionally, we showed how to use SEC to find vector fields of the manifold for points on the manifold, which we use as a local linear space around the data to project to for intermediate points in our on-manifold PGD implementation. 
% In order to obtain a manifold on our dataset we use a synthetic dataset where each class is a single vehicle that is continuously sampled from azimuthal and down look angles. 
% Due to the separability of this dataset the NNs we trained were less susceptible to adversarial attack compared to the results on the CIFAR10 dataset \cite{engstrom_robustness_2019}. 
We were able to successfully obtain on manifold examples that the trained NN misclassifies, showing the promise of on-manifold examples that can be found in input space without reducing down to a latent space. 
% These on-manifold examples were also explained in terms of their intrinsic parameter coordinates, so that the misclassification could become human interpretable. 
Our reported results provided the geometry approximation tool with data that was outside the data used to train the NN classifier, meaning that the output of the on-manifold PGD algorithm would not be a valid input for adversarial regularization.
However, the experiment did provide novel tools for modeling the data manifold in a manner that allowed the on-manifold PGD algorithm to walk in the direction of the NN gradient while remaining on the manifold.
This provided novel examples that were on-manifold but not simply part of the hold out data.
In addition, the Nystr\"{o}m projection onto the manifold provided the intrinsic parameters of the adversarial examples so that the misclassification was human interpretable.
The ability to report the intrinsic parameter of arbitrary points on the manifold opens the door to being able to explain NN decision boundaries in human understandable terms without explicitly sampling all possible inputs.
In non-synthetic data a single class will typically not have continuously varying intrinsic parameters however, so additional work needs to done to transition these tools to real world datasets.

% Some limitations:
% \begin{itemize}
%     \item SEC/CIDM need dense data sampling to perform well, but this increases computational burden 
%     \item In the scenario where we compute geometry independently for each class, we tend to use a small number of classes. Pair this with high data sampling and even basic CNNs perform really well
%     \item One plus is that manifold learning may enable extrapolation off the manifold that NN classifiers usually handle poorly. However, if we only train geometry on the same samples that the network was trained on, this makes things challenging because SEC tends to generalize poorly (at least in our vehicle data dome experiments)
% \end{itemize}

\section*{Conflicts of Interest Statement}
Authors AM and MM are employed by Teledyne Scientific \& Imaging and TS was employed by Teledyne Scientific \& Imaging. The remaining authors declare that the research was conducted int eh absence of any commercial of financial relationships that could be construed as a potential conflict of interest.

\section*{Acknowledgments}
This material is based upon work supported by the Defense Advanced Research Projects Agency (DARPA) under Agreement No. HR00112290079, it has been approved for public release; distribution is unlimited.  
The authors would also like to thank Juan M. Bello-Rivas for many helpful and insightful discussions related to these topics.

\addcontentsline{toc}{section}{References}
\bibliographystyle{plain}
% \bibliography{bibtex/zotero_GoL.bib,bibtex/refs.bib}
\bibliography{bibtex/refs_combo.bib}

\newpage
\appendix

\section{Concise Overview of SEC}\label{sec}

\subsection{Spectral Exterior Calculus (SEC)}\label{sec:secbackground}
% \begin{itemize}
%     \item key insight that motivated SEC 
%     \item outline of theory for decomposition of parallelizable manifolds (frame as nonlinear PCA)
% \end{itemize}

The goal of the SEC is to start with the tools of Diffusion Maps, namely, the eigenfunctions and eigenvalues of the Laplace-Beltrami operator on a manifold, and build the rest of the components of Riemannian geometry.  While there are many aspects of this deep theory, the fundamental ideas can be easily understood by understanding how the SEC lifts the Diffusion Maps representation of functions up to a way of representing vector fields.  Understanding vector fields is the key component of the SEC for the purposes of this project because optimization always involves a vector field, namely the gradient of the objective function.  The SEC starts from the classical understanding in differential geometry that there are two standard ways of thinking about vector fields.  

First, the conventional view of a vector fields is an assignment of an element of $T_xM$ to each $x \in M$.  This representation is given by special kind of map from the manifold $M$ to the tangent bundle $TM = \{(x,v) : x\in M, v \in T_xM\}$ called a \emph{section of the tangent bundle}. To be \emph{section} a $v:M \to TM$ must satisfy $\pi_1(v(x)) = x$.  All this means is that $v(x)$ is an element of the tangent bundle of the form $(x,v_x)$ where $v_x \in T_xM$. So the \emph{section} condition just guarantees that the vector field maps $x$ to a vector in $T_xM$ (rather than the tangent space to some other point).

The second view of vector fields is as linear operators that take $C^{\infty}$ functions to $C^{\infty}$ functions.  In this view, we require, $v:C^{\infty}(M) \to C^{\infty}(M)$ to be linear, $v(a f + g) = a v(f) + v(g)$, and the Liebniz (product) rule, $v(fg)=fv(g)+v(f)g$.  A linear operator that satisfies Liebniz rule is also sometimes called a \emph{derivation}.  

The relationship between the two views is that,
\[ v(f)(x) = v_x \cdot \nabla f(x) = g_x(v_x,\nabla f(x)) = \sum_i (v_x)_i \frac{\partial f}{\partial x_i} \]
where the last equality is the coordinate representation and is independent of the choice of coordinates (the Riemannian metric cancels with the inverse metric built into the definition of the gradient).  Technically speaking, $v_x = \pi_2(v(x))$ in the section of the tangent bundle view.   

\subsubsection{SEC Representations of Vector Fields}\label{SECframe}

The first SEC representation of a vector field is the \emph{frame representation}.  The frame representation writes the vector field as,
\[ v = \sum_{ij} v^{ij} \phi_i \nabla \phi_j \]
where $\phi_i$ are eigenfunctions of the $0$-Laplacian.  This representation is not unique, however there is a natural choice of $v^{ij}$ which is the one that minimizes $\sum_{ij} (v^{ij})^2$.  The SEC returns vector fields in this form, so the eigenfields of the 1-Laplacian are returned as lists of the coefficients $v^{ij}$.  

The second SEC representation of a vector field is the matrix representation of the linear operator in the basis of eigenfunctions of the $0$-Laplacian. 
\[ v_{ij} = \left<\phi_i,v(\phi_j) \right>_{L^2(M)} = \int_M \phi_i(x)v(\phi_j)(x) \, d\textup{vol} = \int_M v_x \cdot \phi_i(x)\nabla\phi_j(x)\, d\textup{vol} = \left<v,\phi_i \nabla \phi_j \right>_{L^2(\Xi)} \]
where $\Xi$ is the space of vector fields on the manifold.  If we are given the coefficients $v_{ij}$ and we want to apply the vector field to a function $f$ we first need to find the generalized Fourier coefficients of $f$ which are given by $\hat f_j = \left<f,\phi_j \right>_{L^2(M)}$ so that $f = \sum_j \hat f_j \phi_j$.  We can then find the generalized Fourier coefficients of $v(f)$ by,
\[ \widehat{v(f)}_i = \left<v(f),\phi_i\right>_{L^2(M)} = \left<v(\sum_j \hat f_j \phi_j),\phi_i\right>_{L^2(M)} = \sum_j \hat f_j \left<v(\phi_j),\phi_i\right>_{L^2(M)} = \sum_j v_{ij}\hat f_j \]
which is the matrix-vector product of the matrix $v_{ij}$ with the vector $\hat f_j$ of generalized Fourier coefficients.  Finally, we can reconstruct the function $v(f)$ from its generalized Fourier coefficients as,
\[ v(f)(x) = \sum_i \widehat{v(f)}_i \phi_i(x) \]

Now, we need a way to move between frame representation and the operator representation.  Suppose someone gives us the frame coefficients, $v^{ij}$, we can then compute the operator coefficients as,
\[ v_{ij} = \left<\phi_i,v(\phi_j) \right>_{L^2(M)} = \left<\phi_i,\sum_{lk} v^{lk}\phi_l\nabla \phi_k(\phi_j) \right>_{L^2(M)} = \sum_{lk} v^{lk}\left<\phi_i, \phi_l\nabla \phi_k \cdot \nabla \phi_j \right>_{L^2(M)} = \sum_{lk} v^{lk}G_{ijlk}\]
where we define,
\[ G_{ijlk} \equiv \left<\phi_i, \phi_l\nabla \phi_k \cdot \nabla \phi_j \right>_{L^2(M)}. \]
So we just need to compute the matrix $G_{ijlk}$ (technically this is a 4-tensor but we think of $i$ and $j$ indexing the rows and $l$ and $k$ indexing the columns of a matrix, so if we use $N$ eigenfunctions, $G$ is an $N^2 \times N^2$ matrix).  $G_{ijlk}$ can be exactly computed using the product formula for the Laplacian 
\begin{align*} \nabla \phi_k \cdot \nabla \phi_j &= \frac{1}{2} \left(\phi_k\Delta\phi_j +\phi_j\Delta\phi_k - \Delta(\phi_j\phi_k) \right) \\
&=\frac{1}{2} \left( \lambda_j\phi_k\phi_j +\lambda_k\phi_j\phi_k - \Delta(\sum_s c_{jks}\phi_s)\right) \\
&=\frac{1}{2} \left( (\lambda_j+\lambda_k)\phi_k\phi_j - \sum_s c_{jks}\Delta(\phi_s)\right) \\
&=\frac{1}{2} \left( (\lambda_j+\lambda_k)\phi_k\phi_j - \sum_s c_{jks}\lambda_s\phi_s \right)\\
&=\frac{1}{2} \left( (\lambda_j+\lambda_k)\sum_{s}c_{jks}\phi_s - \sum_s c_{jks}\lambda_s\phi_s \right)\\
&= \frac{1}{2} \sum_{s}(\lambda_j+\lambda_k -\lambda_s)c_{jks}\phi_s
\end{align*}
where $c_{jks} = \left<\phi_j \phi_k,\phi_s\right>$ are called the \emph{structure constants} of the multiplicative algebra on functions.  Plugging this formula into $G_{ijlk}$ we have,
\[ G_{ijlk} = \frac{1}{2}\sum_{s}(\lambda_j+\lambda_k -\lambda_s)c_{jks} \left<\phi_i, \phi_l \phi_s \right>_{L^2(M)} = \frac{1}{2}\sum_{s}(\lambda_j+\lambda_k -\lambda_s)c_{jks}c_{lsi} \]
so the $G$ matrix can be computed using only the structure constants and the eigenvalues.  This formula is the first truly global closed form representation of the Riemannian metric, which forms the cornerstone of all of Riemannian geometry (if you know the Riemannian metric, you know everything about the geometry of the manifold).  Moreover, we can recover this global representation of the Riemannian metric using only the eigenvalues and eigenfunctions of the Laplace-Beltrami operator, which can be directly estimated from data using diffusion maps.

Of course when we compute this from data we only have finitely many eigenfunctions and eigenvalues, so we have to truncate the summation over $s$.  This formula for $G_{ijlk}$ is the fundamental formula that makes the SEC possible.  Notice that it is building the Grammian matrix on a frame for vector fields using only information from eigenvalues and eigenfunctions of the Laplace-Beltrami operator.  Thus, we have lifted ourselves from talking about functions to being able to describe spaces of vector fields.

\subsubsection{Mapping an SEC Vector Field to Arrows in an Embedding}

Both the frame representation and the operator representation are abstract representation of vector fields.  In order to draw arrows, we need an embedding function $F:M\to\mathbb{R}^n$.  The derivative of the embedding gives a map from the tangent spaces of $M$ into $\mathbb{R}^n$ since,
\[ DF(x) : T_xM \to T_{F(x)}\mathbb{R}^n \cong \mathbb{R}^n. \]
So for a vector $v_x \in T_xM$ we can map this abstract vector to an arrow $DF(x)v_x$.  Now since $F = (F_1,...,F_n)$ where $F_i : M \to \mathbb{R}$ we can write the matrix-vector product,
\[ DF(x)v_x = (DF_1(x)v_x,...,DF_n(x)v_x)^T = (\nabla F_1(x)\cdot v_x,...,\nabla F_n(x)\cdot v_x)^T \]
since $DF_i(x)v_x$ is the directional derivative of $F_i$ in the direction $v_x$, which can also be written as $\nabla F_i(x)\cdot v_x$.  As we saw above, this is the same as applying the operator $v$ to the function $F_i$ and evaluating at $x$, namely,
\[ \nabla F_i(x)\cdot v_x = v(F_i)(x) \]
so that,
\[ DF(x)v_x = (v(F_1)(x),...,v(F_n)(x))^T.  \]
Now we are ready to compute this using the SEC.  As described above, we just need find the generalized Fourier coefficients of $F_1,...,F_n$ and then multiply the vectors of generalized Fourier coefficients by the matrix representation of the vector field $v$.  So ultimately we find that the $k$-th coordinate of the arrow at the point $x$ is given by,
\[ (DF(x)v_x)_k = v(F_k)(x) = \sum_{j} v_{ij}(\hat F_k)_j \phi_i(x) \]

Finally, since the SEC represents vector fields purely in terms of the eigenvectors $\vec\phi_i$, extending these eigenvectors to the entire input spaces immediately extends any vector field represented in the SEC to the entire input space. 

\subsection{Dirichlet Energy on Vector Fields in the SEC Frame}

In the previous section we derived the tensor representation of the Riemannian metric in the frame $\{\phi_i \nabla\phi_j\}_{i,j=1}^{\infty}$ for vector fields.  However, in order to respect the global structure of the manifold we need a well defined energy that we can minimize.  We introduced such an energy in Section \ref{sec:sec_vectors} and it is given by,
\[ E(v,w) \equiv \int_\mathcal{M} (\nabla\times v) \cdot (\nabla\times w) \, d\textup{vol} + \int_\mathcal{M} (\nabla \cdot v)(\nabla \cdot w) \, d\textup{vol}.  \]
This energy comes from the Dirichlet energy on forms in Riemannian geometry which is defined by,
\[ E(\nu,\omega) \equiv \int_\mathcal{M} d\nu \cdot d\omega \, d\textup{vol} + \int_\mathcal{M} (\delta \nu)(\delta \omega) \, d\textup{vol} \]
where $\nu,\omega$ are differential forms, $d$ is the exterior derivatives, and $\delta$ is the co-differential operator. By using the duality between vector fields and differential 1-forms, we were able to define the corresponding energy on vector fields shown above.  The minimizers of the Dirichlet energy on differential 1-forms are the eigenforms of the 1-Laplacian, $\Delta_1 = d\delta + \delta d$, and the dual vector fields of these eigenforms will be minimizers of the energy on vector fields.  Finding these minimizers was a key goal of the SEC \cite{berry_spectral_2020}, and this is possible because we can find a closed formula for the energy on frame elements, $\phi_i\nabla \phi_j$.  First by the product rule for the divergence we have,
\[ \nabla \cdot (\phi_i \nabla \phi_j) = \nabla \phi_i \cdot \nabla \phi_j + \phi_i \nabla \cdot \nabla \phi_j =\nabla \phi_i \cdot \nabla \phi_j - \phi_i \Delta \phi_j =\nabla \phi_i \cdot \nabla \phi_j - \lambda_j \phi_i \phi_j \]
where the negative sign comes from $\Delta$ being the Hodge Laplacian, which on functions is the negative of the Laplace-Beltrami operator, so we have $\Delta f = -\nabla \cdot \nabla f$.
As shown in Section \ref{SECframe}, $\nabla \phi_i \cdot \nabla \phi_j$ can be re-written in terms of the eigenvalues and structure coefficients so that,
\begin{align*} \nabla \cdot (\phi_i \nabla \phi_j) &= \frac{1}{2}\sum_{s}(\lambda_i+\lambda_j -\lambda_s)c_{ijs}\phi_s - \lambda_j \sum_s c_{ijs}\phi_s \\
&=\frac{1}{2}\sum_{s}(\lambda_i-\lambda_j -\lambda_s)c_{ijs}\phi_s.
\end{align*}
This allows us to compute the second intergral in the Dirichlet energy on frame elements as,
\begin{align*}
    \int_\mathcal{M} (\nabla \cdot \phi_i\nabla\phi_j)(\nabla \cdot \phi_k\nabla\phi_l) \, d\textup{vol} &= \left<\nabla \cdot \phi_i\nabla\phi_j,\nabla \cdot \phi_k\nabla\phi_l \right> \\
    &=\frac{1}{4} \left<\sum_{s}(\lambda_i-\lambda_j -\lambda_s)c_{ijs}\phi_s,\sum_{r}(\lambda_k-\lambda_l -\lambda_r)c_{klr}\phi_r \right> \\
    &=\frac{1}{4}\sum_{s,r}(\lambda_i-\lambda_j -\lambda_s)(\lambda_k-\lambda_l -\lambda_r)c_{ijs}c_{klr} \left<\phi_s,\phi_r \right> \\
    &=\frac{1}{4}\sum_{s}(\lambda_i-\lambda_j -\lambda_s)(\lambda_k-\lambda_l -\lambda_s)c_{ijs}c_{kls} 
\end{align*} 
In order to perform a similar calculation on the first integral in the energy, we first need a product rule for the curl operator. While this is straightforward in Euclidean space, on a manifold we require the definition of the cross product of two vector fields, namely set
\[ v\times w \equiv \star(v^\flat \wedge w^\flat)^\sharp \]
then we have the product rule,
\[ \nabla \times (fv) = \star d(fv^\flat)^\sharp = \star (df \wedge v^\flat)^\sharp + f \star (dv^\flat)^\sharp = \nabla f \times v + f\nabla \times v \]
and in particular, since the curl of a gradient is zero, we have 
\[ \nabla \times (\phi_i\nabla\phi_j) = \nabla \phi_i \times \nabla\phi_j \]
since $\nabla \times \nabla \phi_j = 0$.  Finally, this allows us to compute on the frame as,
\begin{align*}
    (\nabla \times (\phi_i\nabla\phi_j)) \cdot (\nabla \times (\phi_k\nabla\phi_l)) &= (\nabla \phi_i \times \nabla\phi_j) \cdot (\nabla \phi_k \times \nabla\phi_l) \\
    &= (\star(d\phi_i \wedge d\phi_j)^\sharp) \cdot (\star(d\phi_k \wedge d\phi_l)^\sharp) \\
    &= (d\phi_i \wedge d\phi_j) \cdot (d\phi_k \wedge d\phi_l) \\
    &= \textup{det}\left[ \begin{array}{cc} \nabla \phi_i\cdot\nabla\phi_k & \nabla \phi_i\cdot\nabla\phi_l \\ \nabla \phi_j\cdot\nabla\phi_k & \nabla \phi_j\cdot\nabla\phi_l  \end{array} \right]\\
    &= (\nabla \phi_i\cdot\nabla\phi_k)(\nabla \phi_j\cdot\nabla\phi_l ) - (\nabla \phi_i\cdot\nabla\phi_l )( \nabla \phi_j\cdot\nabla\phi_k)
\end{align*}
where the third equality follows from both the sharp and star operators being isometries, and the fourth is the definition of the dot product on 2-forms.  This somewhat subtle computation reduces the integrand to a combination of familiar dot products of gradients that are easily represented using eigenvalues and structure constants as,
\[ (\nabla \times (\phi_i\nabla\phi_j)) \cdot (\nabla \times (\phi_k\nabla\phi_l)) = \frac{1}{4} \sum_{s,r}(\lambda_i+\lambda_k -\lambda_s)(\lambda_j+\lambda_l -\lambda_r)c_{iks}c_{jlr}\phi_s\phi_r -(\lambda_i+\lambda_l -\lambda_s)(\lambda_j+\lambda_k -\lambda_r)c_{ils}c_{jkr}\phi_s\phi_r  \]
so that by orthonormality of the eigenfunctions, we have $\int_\mathcal{M}\phi_s\phi_r \, d\textup{vol} =\delta_{sr}$ and the integral becomes,
\[ \int_\mathcal{M} (\nabla \times (\phi_i\nabla\phi_j)) \cdot (\nabla \times (\phi_k\nabla\phi_l)) \, d\textup{vol} = \frac{1}{4}\sum_s (\lambda_i+\lambda_k -\lambda_s)(\lambda_j+\lambda_l -\lambda_s)c_{iks}c_{jls} -(\lambda_i+\lambda_l -\lambda_s)(\lambda_j+\lambda_k -\lambda_s)c_{ils}c_{jks}.\]
Finally, combining the two integrals of the Dirichlet energy on frame elements we have,
\begin{align*}
    E_{ijkl} \equiv E(\phi_i \nabla \phi_j, \phi_k \nabla \phi_l)  &= \frac{1}{4}\sum_s (\lambda_i+\lambda_k -\lambda_s)(\lambda_j+\lambda_l -\lambda_s)c_{iks}c_{jls} -(\lambda_i+\lambda_l -\lambda_s)(\lambda_j+\lambda_k -\lambda_s)c_{ils}c_{jks} \\ &\hspace{30pt}+ (\lambda_i-\lambda_j -\lambda_s)(\lambda_k-\lambda_l -\lambda_s)c_{ijs}c_{kls}.
\end{align*}
While this expression can be condensed a bit (see \cite{berry_spectral_2020}) it is sufficient for computations and for showing the fundamental insight of the SEC which is that these higher energies on vector fields (and tensor fields) can be represented purely in terms of the eigenvalues of the Hodge Laplacian and the structure constants, which come from the eigenfunctions of the Hodge Laplacian.  Since Diffusion Maps and the CIDM are able to estimate these quantities from point clouds, the SEC can compute these energies for point clouds. 

The final component of the SEC is to address the issue of computing the minimizers of these energies when they are represented on a frame instead of a basis.  There is no way around this problem because we are only able to give closed formulas for the energies in the frame, $\{\phi_i\nabla \phi_j\}$, and in general this set will not be a basis.  Thus, once we have constructed the tensors $G_{ijkl}$ and $E_{ijkl}$ representing the inner product and Dirichlet energy respectively, we can reshape these tensors into matrices $G_{(i,j),(k,l)}$ and $E_{(i,j),(k,l)}$ where $(i,j)$ and $(k,l)$ are given unique indices corresponding to those ordered lists.  Thus $G_{(i,j),(k,l)}$ is the inner product of the $(i,j)$-th frame element with the $(k,l)$-th frame element. Since the fundamental difference between a frame and a basis is that the frame elements are not independent, there are non-trivial linear combinations of frame elements that sum to zero, and so we should expect the $G$ matrix to have zero eigenvalues, although due to truncation it may only have very small eigenvalues.  

Using the matrices, $G$ and $E$, we would like to find vectors of coefficients, $\vec c$ that minimize,
\[ \frac{\vec c^\top E \vec c}{\vec c^\top G \vec c} \]
which corresponds to solving the generalized eigenvalue problem $E\vec c = \eta G\vec c$.  These vectors of coefficients will then define a linear combination of frame elements $\sum_{i,j} c_{(i,j)}\phi_i\nabla\phi_j$ that defines a vector field with minimal Dirichlet energy subject to a norm constraint. However, because we expect $G$ to be numerically rank deficient, this is not a well defined problem on a frame.  Instead, we need to first move to a basis by removing the redundant frame representations.  It is critical to do this using the inner product which defines the natural domain of the 1-Laplacian, namely the Sobolev inner product which is represented by the matrix $E+G$.  Thus, the first step is to compute an eigendecomposition of the matrix $E+G = USU^\top$ and then to select a threshold of the eigenvalues to isolate the redundant combinations.  In practice we have found that the threshold should be set at a fraction of the leading eigenvalue, e.g. $\tau = 10^{-3}S(1,1)$.  Once a threshold has been selected, let $\tilde U$ be the columns of $U$ whose corresponding eigenvalues are larger than $\tau$.  These columns represent a well conditioned basis relative to the Sobolev inner product.  We can now project the eigenvalue problem down to this basis by constructing $\tilde E = \tilde U^\top E \tilde U$ and $\tilde G = \tilde U^\top G \tilde U$.  Now we can solve the eigenvalue problem $\tilde E \tilde c = \eta \tilde G \tilde c$ which is well defined in the basis, and finally we can recover the frame coefficients $\vec c = \tilde U \tilde c$.  

Using $\vec c$ we can reconstruct the vector field by $\sum_{i,j} c_{(i,j)}\phi_i\nabla\phi_j$, and then we can use the SEC tools discussed above to visualize or compute arrows in embedding spaces, such as the data space.  By using several of the eigenvectors, $\tilde c$, with the smallest eigenvalues, we can construct several vector fields that together will span the tangent space of the manifold.  The key advantage of tangent spaces approximated in this manner, are that the vector fields constructed from the SEC are seeing the global structure of the data set, unlike a local PCA approach that can only see nearby neighbors.

\end{document}